\documentclass{article}

\usepackage{amsmath}
\usepackage{amssymb}
\usepackage{mathtools}
\usepackage{amsthm}
\usepackage{algorithm}
\usepackage{algorithmic} 
\usepackage{framed}
\usepackage{url}
\usepackage{graphicx}
\usepackage{booktabs}
\usepackage{makecell}

\usepackage{microtype}
\usepackage{graphicx}
\usepackage{subcaption}
\usepackage{xspace}

\usepackage[accepted]{icml2026}

\usepackage{hyperref}

\usepackage{chngcntr}

\usepackage[capitalize,noabbrev]{cleveref}

\theoremstyle{plain}
\newtheorem{theorem}{Theorem}[section]

\theoremstyle{definition}
\newtheorem{definition}[theorem]{Definition}

\theoremstyle{remark}

\usepackage[textsize=tiny]{todonotes}

\newcommand{\framework}{\texttt{MetaEns}\xspace}

\begin{document}

\twocolumn[
\icmltitle{Automatic Unsupervised Ensemble Outlier Model Selection--Extended Version}

\icmlsetsymbol{equal}{*}

\begin{icmlauthorlist}
\icmlauthor{Hong-Phuc Phan}{equal,fpt}
\icmlauthor{Tuan-Anh Vu}{equal,ctut}
\icmlauthor{Tung Kieu}{equal,aau}
\icmlauthor{Son Ha Xuan}{rmit}
\icmlauthor{Bin Yang}{ecnu}
\icmlauthor{Christian S. Jensen}{aau}
\end{icmlauthorlist}

\icmlaffiliation{fpt}{Department of Software Engineering, FPT University, Vietnam}
\icmlaffiliation{ctut}{Department of Information Technology, Can Tho University of Technology, Vietnam}
\icmlaffiliation{rmit}{School of Business, RMIT University, Vietnam}
\icmlaffiliation{aau}{Department of Computer Science, Aalborg University, Denmark}
\icmlaffiliation{ecnu}{School of Data Science and Engineering,
East China Normal University, China}

\icmlcorrespondingauthor{Tung Kieu}{tungkvt@cs.aau.dk}
\icmlcorrespondingauthor{Son Ha Xuan}{Ha.Son@rmit.edu.vn}

\icmlkeywords{Anomaly Detection, Model Selection, Data Mining, Unsupervised Learning}

\vskip 0.3in
]

\printAffiliationsAndNotice{\icmlEqualContribution}

\begin{abstract}

Unsupervised outlier detection is attractive because it eliminates the need for labeled data. Moreover, forming multi-model ensembles can improve detection robustness. However, composing an ensemble without labeled data is challenging. Naively composed ensembles can suffer from ensemble saturation, where redundant or unreliable detection models degrade performance and incur unnecessary computation. We propose \framework, an automatic unsupervised framework for selecting ensembles of outlier detection models. Using labeled meta-datasets, \framework learns a model that predicts marginal ensemble gains, estimating the expected improvement from adding a candidate model to a partially constructed ensemble. At test time, this learned signal is combined with a submodular-inspired proxy objective that enforces diminishing returns through diversity-aware discounting and family-level risk regularization, thereby enabling greedy sequential selection with adaptive early stopping. As a result, \framework constructs compact, high-quality ensembles without access to ground-truth labels. Experiments on 39 real-world datasets show that \framework consistently outperforms state-of-the-art unsupervised selectors and ensemble baselines, achieving higher average precision while using fewer models.

\end{abstract}
\section{Introduction}
\label{sec:introduction}

Outlier detection plays a critical role in applications like fraud detection, network security, medical diagnosis, and system monitoring~\cite{DBLP:journals/csur/ChandolaBK09,DBLP:journals/pieee/RuffKVMSKDM21}. However, in many real-world scenarios where ground-truth labels are unavailable, and outlier detection must be performed in a fully unsupervised manner~\cite{DBLP:journals/sigkdd/ZimekCS13}. Further, contamination rates are often unknown, and data distributions vary widely across tasks~\cite{DBLP:conf/nips/HanHHJ022,DBLP:conf/icde/ZhangKQGHZJY26}. These factors make it difficult to not only detect anomalies but also evaluate and compare detection models.


Existing studies have proposed diverse unsupervised outlier detectors based on density estimation~\cite{DBLP:conf/sigmod/BreunigKNS00}, isolation mechanisms~\cite{DBLP:conf/icdm/LiuTZ08}, reconstruction errors~\cite{DBLP:conf/iclr/ZongSMCLCC18}, and deep representations~\cite{DBLP:conf/icml/RuffGDSVBMK18}. Despite their success in specific settings, no single detector performs reliably across diverse datasets. This observation has motivated the use of ensemble methods, which aim to improve robustness by aggregating multiple detectors~\cite{DBLP:journals/pvldb/CamposKGHZYJ21,DBLP:conf/ijcai/KieuYGJ19}. In supervised learning, ensembles can be trained and validated using labeled data. In contrast, constructing effective ensembles for unsupervised outlier detection remains an open challenge \cite{DBLP:journals/sigkdd/ZimekCS13}.

A key challenge is to select models when no labels are available~\cite{DBLP:journals/tkdd/MarquesCSZ20}. Without ground-truth feedback, it is unclear which detection models are reliable on a given dataset or whether adding a new detector will improve or degrade ensemble performance~\cite{DBLP:conf/icdm/RayanaZA16,DBLP:journals/sigkdd/AggarwalS15}. As a result, many unsupervised ensemble methods rely on fixed aggregation strategies, such as averaging scores from all available detectors or selecting a fixed number of top-ranked detection models~\cite{DBLP:conf/sdm/ZhaoNHL19}. Methods employing these strategies suffer from \emph{ensemble saturation}: beyond a small ensemble size, adding more detectors yields diminishing or even negative returns due to redundancy, conflicting rankings, or systematically poor models. Moreover, fixed-size ensembles are inherently inflexible and cannot adapt to dataset-specific complexity.


Recent unsupervised model selection methods attempt to address these challenges by leveraging meta-learning across labeled auxiliary datasets~\cite{DBLP:journals/pami/HospedalesAMS22}. Notably, frameworks such as \texttt{MetaOD}~\cite{DBLP:conf/nips/ZhaoRA21} and \texttt{ELECT}~\cite{DBLP:conf/icdm/0016ZA22} learn to recommend a single detector for a new unlabeled task based on task similarity or historical performance patterns. While effective, these methods are limited to singleton selection and do not address the more general problem of adaptive ensemble construction, where multiple complementary detectors may be required to capture diverse outlier patterns~\cite{DBLP:conf/aaai/ChengWL020}.

We address this gap by formulating unsupervised ensemble outlier model selection as a sequential decision problem. Our key insight is that although the true marginal benefit of adding a detector to an ensemble is unobservable at test time, its structure can be learned offline from labeled meta-datasets. Building on this idea, we propose \framework, a framework that learns to predict the marginal ensemble gain of candidate detectors conditioned on the current ensemble state. At inference time, \framework greedily constructs an ensemble by maximizing a submodular-inspired proxy objective \cite{DBLP:journals/mp/NemhauserWF78} that integrates the predicted gain with explicit mechanisms for diversity control and risk mitigation.

Specifically, \framework introduces two principles that are crucial for unsupervised ensemble construction. First, we enforce \emph{diminishing returns} through similarity-based discounting that penalizes candidates that introduce redundancy with already selected detectors \cite{DBLP:journals/ftml/KuleszaT12}. Second, we incorporate \emph{family-risk regularization}, which discourages the selection of multiple detectors from algorithmic families with a history of poor or unstable performance. Together, these components yield a proxy objective that favors compact, diverse ensembles and naturally supports adaptive early stopping when no further improvement is expected.

We evaluate \framework on a benchmark of 39 real-world anomaly detection datasets \cite{DBLP:conf/nips/HanHHJ022} using a large pool of 297 candidate detectors spanning multiple algorithmic families. Extensive experiments show that \framework is able to consistently outperform strong unsupervised baselines and recent meta-learning approaches. Notably, \framework achieves higher detection accuracy while selecting fewer models than fixed-size ensembles, and it exhibits strong resilience by recovering performance even when an initially selected detector performs poorly.

In summary, our contributions are threefold:
\begin{itemize}
    \item We formulate the problem of unsupervised ensemble outlier model selection and cast it as a sequential decision process without access to labels.
    \item We propose \framework, a meta-learning framework that predicts marginal ensemble gains and combines these with a diversity- and risk-aware proxy objective to enable adaptive ensemble construction with early stopping.
    \item We provide experimental results across 39 datasets, supported by ablation studies, showing that compact, adaptively sized ensembles can outperform larger fixed ensembles in fully unsupervised settings.
\end{itemize}

\section{Preliminaries}
\label{sec:preliminaries}

\begin{definition}[Dataset]
    A dataset $\mathbf{X}$ is a finite collection of data instances (or data points) $\mathbf{X} = \{\mathbf{x}_{1}, \mathbf{x}_{2}, \ldots, \mathbf{x}_{N}\}$, where each instance $\mathbf{x}_{i} \in \mathbb{R}^{d}$ is a $d$-vector. We denote $|\mathbf{X}| = N$ as the cardinality of the dataset.
\end{definition}



\begin{definition}[Unsupervised Outlier Detection]
Given a dataset $\mathbf{X} \in \mathbb{R}^{N \times d}$, an unsupervised outlier detection model, or detector, learns a scoring function 
$f: \mathbf{X} \rightarrow \mathbb{R}^{N}$ that assigns an outlier score $o_i = f(\mathbf{x}_i)$ to each instance $\mathbf{x}_i \in \mathbf{X}$, where larger values indicate a higher likelihood of being an anomaly. 
A decision function can be derived by thresholding the scores at a user-defined level $\tau$, yielding outlier labels 
$\hat{y}_i = \mathbb{I}(o_i > \tau)$. 
The resulting outlier set is defined as follows.
\[
\mathbf{O} = \{ \mathbf{x}_i \in \mathbf{X} \mid \hat{y}_i = 1 \}
\]
\end{definition}

\begin{definition}[Unsupervised Outlier Model Selection]
    Let $\Omega = \{f_{1}, f_{2}, \ldots, f_{K}\}$ be the set of $K$ candidate outlier detection models. Each $f_{i} \in \Omega$ can be seen as a $(\mathtt{detector}, \mathtt{configuration})$ tuple, where the $\mathtt{configuration}$ denotes a set of hyperparameters of the $\mathtt{detector}$. Let $\mathbf{X} = \{\mathbf{x}_{1}, \mathbf{x}_{2}, \ldots, \mathbf{x}_{N}\}$ be an unlabeled dataset. Each outlier detection model $f_{i}$ acts as an outlier scoring function $f_{i}: \mathbb{R}^{N \times d} \mapsto \mathbb{R}^{N}$, which assigns an outlier score to each instance in $\mathbf{X}$. The task of unsupervised outlier model selection is then to choose the model $f^{*}$ as follows. $$f^{*} = \arg\min_{f_{i} \in \Omega} \Gamma(f_{i}, \mathbf{X})$$ 
    \noindent Here, $\Gamma(\cdot)$ is an unsupervised evaluation criterion that estimates the quality of model $f_{i}$ based solely on the distributional properties of its output scores without using labeled outlier or normal instances.
\end{definition}

\begin{framed}
    \noindent\textbf{Problem Definition: Unsupervised Ensemble Outlier Model Selection.}   
    Let $\Omega = \{f_{1}, f_{2}, \ldots, f_{K}\}$ be a set of $K$ candidate outlier detection models. Each model $f_{i} \in \Omega$ can be seen as a $(\mathtt{detector}, \mathtt{configuration})$ tuple, where the $\mathtt{configuration}$ denotes a specific set of hyperparameters of the $\mathtt{detector}$. Let $\mathbf{X} = \{\mathbf{x}_{1}, \mathbf{x}_{2}, \ldots, \mathbf{x}_{N}\}$ be an unlabeled dataset. Each outlier detection model $f_{i} \in \Omega$ acts as an outlier scoring function $f_{i}: \mathbb{R}^{N \times d} \mapsto \mathbb{R}^{N}$ that assigns an outlier score to each instance in $\mathbf{X}$. A candidate ensemble $P \subseteq \Omega$ is evaluated as a whole, not as an independent sum of member-model criteria. Specifically, we first aggregate member scores into a single ensemble score vector,
    $$\mathbf{o}_{P} = \frac{1}{|P|}\sum_{f \in P} f(\mathbf{X}).$$
    The task of unsupervised ensemble outlier model selection is to choose a set of models $P \subseteq \Omega$ as follows.
    $$P^{*} = \arg\max_{P \subseteq \Omega} \psi(P;\mathbf{X}) \quad \text{s.t.} \quad |P| \le \eta.$$ 
    \noindent Here, the criterion $\psi(P;\mathbf{X})$ estimates the quality of the aggregated ensemble score vector $\mathbf{o}_{P}$ without using any ground-truth labels, and $\eta$ is a budget on the ensemble size.
\end{framed}

\begin{definition}[Meta-datasets]
    We assume access to a collection of labeled meta-datasets $\mathcal{M} = \{(\mathbf{M}_{1}, \mathbf{y}_{1}), (\mathbf{M}_{2}, \mathbf{y}_{2}), \ldots, (\mathbf{M}_{L}, \mathbf{y}_{L})\}$ where each $\mathbf{M}_{i} \in \mathbb{R}^{N_{M_{i}} \times d_{M_{i}}}$ is a dataset and $\mathbf{y}_{i} \in \{0, 1\}^{N_{M_{i}}}$ provides ground-truth anomaly labels. These meta-datasets are used exclusively for evaluating and selecting candidate outlier detection models. At test time, we are given an unlabeled dataset $\mathbf{X} = \{\mathbf{x}_{1}, \mathbf{x}_{2}, \ldots, \mathbf{x}_{N}\}$ that satisfies $\mathbf{X} \sim \mathbb{P}_{\mathbf{X}}$ and $\mathbb{P}_{\mathbf{X}} \neq \mathbb{P}_{\mathbf{M}_{i}}, \forall i$, i.e., the test distribution differs from the distributions underlying the meta-datasets.
\end{definition}
\section{Methodology}
\label{sec:methodology}

\begin{figure*}[t]
\centering
\includegraphics[width=\textwidth]{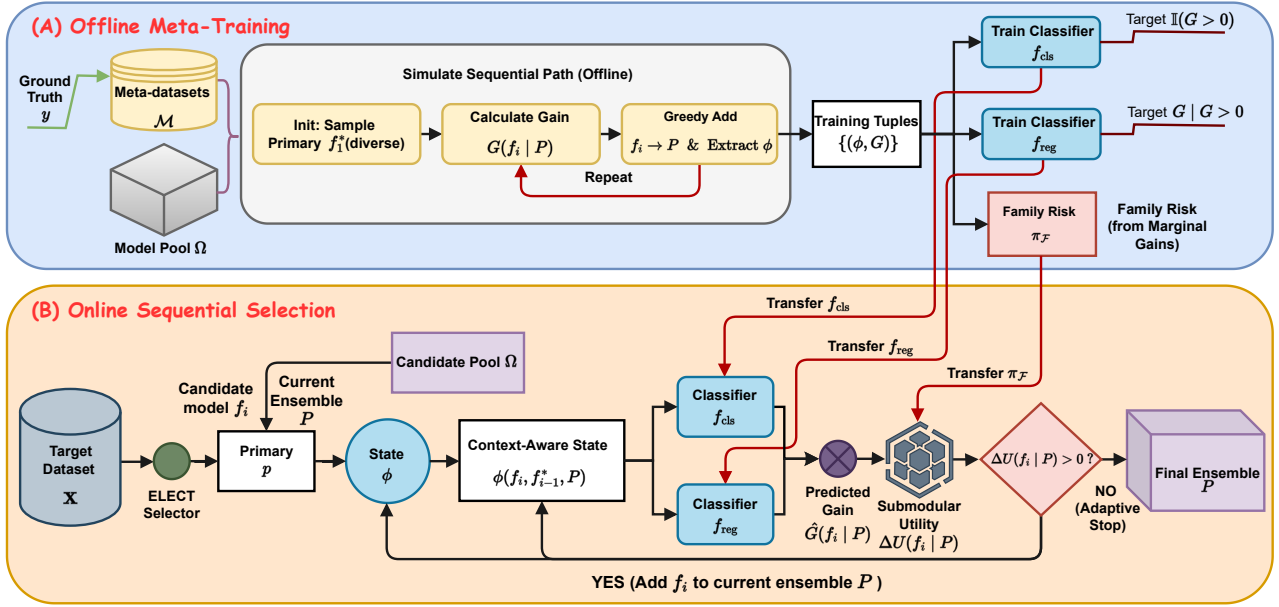}
\caption{Overview of \framework. \textbf{(A) Offline Meta-Training:}
Oracle-greedy rollouts on labeled meta-datasets $\mathcal{M}$ generate
state--gain pairs for partial ensembles. These pairs supervise a two-part gain
predictor consisting of classifier $f_{\text{cls}}$, which estimates whether a
candidate will improve the ensemble, and regressor $f_{\text{reg}}$, which
estimates the positive gain magnitude. Family-risk priors $\pi_{\mathcal{F}}$
are computed from lower-tail oracle gains to identify detector families with
historically harmful additions. \textbf{(B) Online Selection:} On an unlabeled
target dataset, \framework starts from a primary detector and greedily expands
the ensemble by maximizing a proxy utility that combines predicted gain,
redundancy discounting $\gamma$, and family-risk penalty. Selection stops
adaptively when no remaining candidate has positive utility, yielding a compact
label-free ensemble.}
\label{fig:overview}
\end{figure*}

\subsection{Framework Overview}
\framework encompasses two phases: \emph{offline meta-training} and \emph{online model selection}.

\textbf{Offline meta-training.}
Given a collection of labeled meta-datasets, we simulate sequential ensemble construction and compute the \emph{true} marginal gain of adding a candidate detector to a partial ensemble using ground-truth labels. These oracle gains provide supervision for learning a meta-model that predicts the \emph{expected} marginal gain conditioned on the current ensemble context.

\textbf{Online model selection.}
At test time, \framework is applied to a new unlabeled dataset. The procedure starts by selecting a high-quality \emph{primary} detector that serves as an anchor. \framework then expands the ensemble by greedily adding detectors with high predicted utility. Since the predicted gain is a learned one and may be noisy, we introduce a \emph{submodular-inspired} proxy objective that (i) discounts redundant candidates to induce diminishing returns and (ii) penalizes risky algorithm families based on meta-training history. This proxy enables \emph{adaptive early stopping}: selection terminates once no candidate is expected to provide positive utility.

Fig.~\ref{fig:overview} summarizes the interaction between offline meta-training and online selection.

\subsection{Offline Meta-training}

\paragraph{Algorithmic families.}
We partition the candidate pool $\Omega$ into a fixed set of \emph{algorithmic
families} $\mathcal{F}$, where each family groups detectors sharing the same
underlying algorithmic paradigm. For instance, all \texttt{LOF} variants
(differing only in $k$ or distance metric) belong to the same family, as do
all \texttt{IForest} variants (differing in number of estimators or subsampling
ratio). The 297-model pool spans 8 base families: \texttt{IForest},
\texttt{LOF}, \texttt{kNN}, \texttt{HBOS}, \texttt{OCSVM}, \texttt{LODA},
\texttt{ABOD}, and \texttt{COF}. Family assignments are fixed from metadata
and do not change across datasets or training steps. We use a single
family-risk percentile, the 10th percentile of oracle marginal gains, for all
families so that a family is penalized only when its worst-case historical
contributions are systematically negative. This percentile is fixed globally using only
meta-training trajectories and is never tuned on the target dataset. It is a
conservative lower-tail statistic: lower percentiles may miss families with
occasional catastrophic failures, whereas higher percentiles can over-penalize
average families that still provide useful orthogonal diversity. We evaluate
the sensitivity to this family definition in App.~\ref{app:subsec:family_ablation}
and the sensitivity to the risk percentile in App.~\ref{app:subsec:hyperparams}.

Let $\mathcal{M}=\{(\mathbf{M}_\ell,\mathbf{y}_\ell)\}_{\ell=1}^{L}$ be the labeled meta-datasets, and let $\Omega=\{f_1,\dots,f_K\}$ denote the candidate detector pool (detector, configuration). For a dataset $(\mathbf{M},\mathbf{y})\in\mathcal{M}$ and an ensemble (model set) $P\subseteq\Omega$, we define the ensemble score vector as the mean of the member scores:
\begin{equation}
\mathbf{o}_{P} \;=\; \frac{1}{|P|}\sum_{f \in P} f(\mathbf{M}),
\label{eq:outlier_score_set}
\end{equation}
where $f(\mathbf{M})\in\mathbb{R}^{|\mathbf{M}|}$ is the outlier-score vector produced by $f$ on $\mathbf{M}$.
We use the mean as the default aggregation rule because it preserves continuous score information from all selected detectors while reducing sensitivity to any single poorly calibrated model. In App.~\ref{app:subsec:score_combiner_ablation}, we compare this choice against median, max, and min aggregation and find that mean aggregation provides the strongest and most stable performance.

\paragraph{Oracle marginal gain.}
We define the \emph{true} marginal gain of adding $f_i\in \Omega\setminus P$ to $P$ as the improvement in Average Precision:
\begin{equation}
G(f_{i}\mid P) \;=\; \mathrm{AP}(\mathbf{o}_{P \cup \{f_i\}}, \mathbf{y}) \;-\; \mathrm{AP}(\mathbf{o}_{P}, \mathbf{y}),
\label{eq:true_marginal_gain}
\end{equation}
where $\mathrm{AP}(\cdot,\cdot)$ is computed from the ensemble score vector and the ground-truth label vector. We use AP as it is threshold-independent and robust under severe class imbalance, which is common in outlier detection.

\paragraph{Generating training trajectories.}
To obtain informative supervision, we construct meta-training trajectories using an \emph{oracle greedy policy}. For each meta-dataset, we initialize the ensemble with a primary model $f^{*}_{1}$ chosen to maximize $\mathrm{AP}(f(\mathbf{M}),\mathbf{y})$. We then iteratively add models that maximize the true gain in Eq.~\ref{eq:true_marginal_gain}. This strategy exposes the meta-model to high-quality partial ensembles, avoiding training dominated by arbitrary or low-signal states.

We empirically verify that alternative strategies such as $\varepsilon$-greedy 
exploration or reduced meta-dataset sizes consistently underperform oracle 
greedy rollouts, confirming that aligning training with test-time behavior 
is both sufficient and sample-efficient (see App.~\ref{app:subsec:trajectory_ablation}).

At each step $i$, for every candidate $f\in\Omega\setminus P$, we compute a state representation $\phi(f, f^{*}_{i-1}, P)$ and its oracle gain $G(f\mid P)$, obtaining supervised pairs $\{(\phi(\cdot),G(\cdot))\}$ across multiple ensemble sizes and datasets. These pairs are used to learn a predictor of marginal gains.

\paragraph{State representation.}
The state representation $\phi(f_{i}, f^{*}_{i-1}, P)$ encodes three factors essential to sequential ensemble construction: (i) \emph{redundancy} between the candidate and the most recent selection, (ii) \emph{compatibility} between the candidate and the current ensemble, and (iii) the \emph{selection stage} via $|P|$.
We deliberately restrict \framework to score-level summaries rather than raw inputs, learned embeddings, or detector internals. This score-only design makes the representation independent of the original data dimensionality and avoids assumptions about whether a detector is classical, neural, image-based, or text-based. As a result, the same meta-feature map can be applied across heterogeneous detector pools and modality-transfer settings, which we evaluate empirically in Section~\ref{sec:experiments}.

We compute a base feature extractor on normalized score vectors that includes correlation measures (e.g., Spearman correlation, cosine similarity), distributional statistics (e.g., entropy, kurtosis), and overlap-based agreement (e.g., Jaccard similarity over top-ranked instances; see App.~\ref{app:subsec:features} for full definitions). Concretely, we compute:
(i) $\phi_{f_i,f^{*}_{i-1}}\in\mathbb{R}^{d_{\text{base}}}$ from $(\mathbf{o}_{f_i},\mathbf{o}_{f^{*}_{i-1}})$,
(ii) $\phi_{f_i,P}\in\mathbb{R}^{d_{\text{base}}}$ from $(\mathbf{o}_{f_i},\mathbf{o}_{P})$, and
(iii) $\phi_{f^{*}_{i-1},P}\in\mathbb{R}^{d_{\text{base}}}$ from $(\mathbf{o}_{f^{*}_{i-1}},\mathbf{o}_{P})$.
The final state is:
\begin{equation}
\phi(f_{i},f^{*}_{i-1},P) = (\phi_{f_{i},f^{*}_{i-1}}, \phi_{f_{i},P}, \phi_{f^{*}_{i-1},P}, |P|).
\label{eq:state_rep}
\end{equation}

\paragraph{Marginal gain modeling.}
Positive marginal gains become increasingly sparse as the ensemble grows: once a strong partial ensemble has been formed, most remaining candidates are redundant or harmful and therefore have $G(f_i\mid P)\le 0$. The resulting gain distribution is zero-inflated, with many non-positive targets and a small tail of genuinely useful additions. A single regressor trained on this distribution can minimize squared error by predicting small positive values for many candidates, which is undesirable because it can still trigger the selection of redundant models. We therefore use a two-part hurdle-style model that separates \emph{whether} a candidate improves the ensemble from \emph{how much} it improves the ensemble when improvement occurs:
\begin{equation}
\hat{G}(f_{i}\mid P) = f_{\text{cls}}(f_{i}\mid P)\cdot f_{\text{reg}}(f_{i}\mid P),
\label{eq:two_part}
\end{equation}
where $f_{\text{cls}}(f_{i}\mid P)=\mathbb{P}(G(f_i\mid P)>0)$ predicts the probability of improvement and $f_{\text{reg}}(f_{i}\mid P)=\mathbb{E}[G(f_i\mid P)\mid G(f_i\mid P)>0]$ predicts the gain magnitude conditioned on being positive. This decomposition lets the classifier act as a gate for sparse positive gains, while the regressor focuses only on the scale of useful additions.

For each training pair $(\phi, G)$, the classification target is $y_{\text{cls}}=\mathbb{I}(G>0)$. The regression target is $y_{\text{reg}}=\max(0,G)$, optimized only on samples with $G>0$. This design is validated by the ablation in Table~\ref{tab:ablation}: replacing the two-part architecture with a single gain predictor reduces AP from $0.4308$ to $0.4133$ and worsens the average rank from $59.3$ to $87$.

\paragraph{Why ExtraTrees?}
We instantiate both $f_{\text{cls}}$ and $f_{\text{reg}}$ with \texttt{ExtraTrees}~\cite{DBLP:journals/ml/GeurtsEW06}. A single decision tree has high variance and can overfit the moderate-size meta-training corpus, especially because our state representation contains dense, nonlinear score-statistics derived from pairwise detector behavior. ExtraTrees reduces this variance through an ensemble of randomized trees; compared with Random Forests, its more randomized split thresholds provide stronger regularization, and compared with boosting models such as \texttt{XGBoost} or \texttt{LightGBM}, its bagging-style variance reduction better matches our goal of transferring to unseen anomaly-detection datasets rather than fitting residual errors on the meta-training corpus. Empirically, we compare six meta-model families in App.~\ref{app:subsec:meta_model_analysis}; \texttt{ExtraTrees} achieves the best AP ($0.4308\pm0.0064$) and average rank ($59.3\pm6.96$), outperforming alternatives such as \texttt{XGBoost}, \texttt{Random Forest}, and multilayer perceptrons while retaining competitive training and inference costs.

\subsection{Online Model Selection}
At test time, the true gain $G(\cdot)$ is unavailable. We therefore select ensembles using a proxy objective that combines the predicted gain $\hat{G}$ with explicit redundancy and risk control. Our goal is to induce \textit{diminishing returns} behavior---a hallmark of submodular maximization---without requiring the learned predictor to satisfy submodularity.

\paragraph{Proxy marginal utility.}
Given a current ensemble $P$ (with the last selected model being $f^{*}_{i-1}$), the marginal utility of adding candidate $f_i$ is:
\begin{equation}
    \Delta U(f_{i}\mid P) = \gamma(f_{i},P)\cdot \Big(\hat{G}(f_{i}\mid P) - \lambda_{\text{fam}}\pi_{\mathcal{F}(f_{i})}\Big),
    \label{eq:marginal-utility}
\end{equation}
where $\gamma(f_i,P)\in(0,1]$ discounts redundant candidates, and $\lambda_{\text{fam}}\pi_{\mathcal{F}(f_i)}\ge 0$ penalizes families with historically negative tail behavior. The hyperparameters $\beta$ and $\lambda_{\text{fam}}$ control the strength of redundancy discounting and family-risk regularization.

\paragraph{Redundancy discount.}
We define the discount factor as
\begin{equation}
\gamma(f_{i},P) = \frac{1}{1 + \beta\cdot \operatorname{sim}_{\max}(f_{i},P)},
\label{eq:discount-factor}
\end{equation}
where $\operatorname{sim}_{\max}(f_i,P)$ is the maximum similarity between $f_i$ and any model in $P$:
\begin{equation}
\operatorname{sim}_{\max}(f_{i},P) =
\begin{cases}
\max\limits_{f_{j} \in P} \operatorname{sim}(f_{i}, f_{j}) & \text{if } |P| > 0,\\
0 & \text{if } P = \emptyset,
\end{cases}
\label{eq:sim-max}
\end{equation}
where $\operatorname{sim}(f_i,f_j)$ is the Jaccard similarity between the sets of the top-$k_{\text{top}}$ ranked instances of $f_i$ and $f_j$. We use the maximum (rather than average) similarity to prevent selecting near-duplicates of any existing ensemble member, which is particularly harmful without validation labels.

The discount $\gamma(\cdot)$ induces diminishing returns in the following sense: since $\operatorname{sim}_{\max}(f_i,P\cup\{f\})\ge \operatorname{sim}_{\max}(f_i,P)$ for any added model $f$, the redundancy penalty is non-decreasing with ensemble growth, dampening the marginal utility of redundant candidates. We therefore treat the objective as submodular-inspired rather than formally submodular: a strict guarantee would require the learned black-box gain predictor $\hat{G}(f_i\mid P)$ itself to satisfy diminishing returns for all detector subsets, which is difficult to enforce without constraining the predictor and potentially degrading marginal-gain accuracy. Our proxy instead preserves the practical benefits needed here---greedy efficiency, redundancy control, and natural early stopping---while allowing flexible learned interactions among heterogeneous detectors.

\paragraph{Family-risk regularization.}
To reduce downside risk, we introduce a family-level prior that penalizes candidates from algorithmic families with negative lower-tail historical gains. Let $\mathcal{F}:\Omega\to\mathcal{F}$ map each candidate to an algorithm family. For a family $F\in\mathcal{F}$, we define the family risk as the 10th percentile of oracle gains observed during meta-training:
\begin{equation}
\text{Risk}_{F}(\mathcal{M}) = Q_{0.10}\left(\{G(f \mid P) \mid f \in \mathcal{M}_{F},\; |P| \geq 1\}\right),
\label{eq:family-risk}
\end{equation}
and convert it into a non-negative penalty:
\begin{equation}
\pi_{F} = \max\left(0, -\text{Risk}_{F}(\mathcal{M})\right)
\label{eq:family-penalty}
\end{equation}
Here, $\mathcal{M}_{F}$ collects candidates in family $F$ across all meta-datasets and all ensemble states encountered during meta-training. Intuitively, if a family occasionally produces strongly negative marginal gains, it receives a higher penalty. This is to avoid harmful additions in the absence of labels.
For previously unseen families ($\mathcal{M}_{F}=\emptyset$), we set $\text{Risk}_{F}(\mathcal{M})=0$ and thus $\pi_F=0$, treating them as neutral rather than risky. This avoids injecting a manual penalty against novel detector families before observing any negative evidence, and keeps the framework extensible to zero-shot use of future detectors. In practice, all models in our main pool belong to known families, so this rule mainly affects plug-and-play extensions.

\subsection{Framework Algorithm}
\label{subsec:algorithm}

The detailed framework algorithm is provided in App.~\ref{app:subsec:algorithm}, where we present the offline meta-training and online model selection procedure in pseudocode form.

\subsection{Complexity Analysis}
\label{subsec:complexity_analysis}

We conduct a computational complexity analysis to clarify how the proposed framework scales with the size of the candidate model pool and dataset size. The detailed computational complexity analysis is given in App.~\ref{app:subsec:complexity_analysis}.

\section{Experiments}
\label{sec:experiments}

\subsection{Experimental Settings}

\paragraph{Datasets.}
We use the 39-dataset benchmark introduced in \texttt{ELECT}~\cite{DBLP:conf/icdm/0016ZA22},
which consists of independent, real-world tabular datasets drawn from the
Outlier Detection Data Sets (ODDS)\footnote{https://giftpathao.com} and UCI Machine Learning Repository~\cite{DBLP:journals/datamine/CamposZSCMSAH16} repositories. The datasets vary substantially, with the number of samples $n$ ranging from 129 to 49,097 (median $n = 1456$) and the dimensionality $d$ ranging from 4 to 400 (median $d = 21$). Contamination rates range from $0.03\%$ to $35\%$ (median: $2.3\%$). The collection covers three data-type categories: numeric-only (18 datasets), categorical-only (2 datasets), and mixed-type (19 datasets). Details of the datasets are provided in App.~\ref{app:subsec:datasets}.

\paragraph{Candidate Model Pool.}
We construct a large candidate pool of 297 unsupervised outlier detection models spanning 8 widely-used algorithmic families: Isolation Forest (\texttt{IForest}), Local Outlier Factor (\texttt{LOF}), k-Nearest Neighbors (\texttt{kNN}), Histogram-based Outlier Score (\texttt{HBOS}), One-Class SVM (\texttt{OCSVM}), \texttt{LODA}, \texttt{ABOD}, and \texttt{COF}. This diverse pool is generated by systematically varying hyperparameters within each family. For instance, we vary $k$ among $\{5, 10, 15, \dots, 100\}$ for \texttt{kNN} and \texttt{LOF}, contamination factors for \texttt{IForest}, and kernel types for \texttt{OCSVM}. The full specification of the algorithms and hyperparameter grids is detailed in App.~\ref{app:subsec:model_pool}.


\paragraph{Baselines.}
We compare \framework against 19 unsupervised baselines and a supervised greedy oracle upper bound, spanning standard detectors, deep learning methods, and meta-learning or ensemble-based selectors. For ensemble baselines that depend on an ensemble size parameter $k$, we report the best-performing $k$ (selected to maximize average AP), giving these fixed-size baselines an oracle advantage.

The baselines include standard methods such as random selection (\texttt{Singleton}), classical detectors (e.g., \texttt{LOF}); na\"ive and random ensemble methods (e.g., \texttt{IForest}, \texttt{Random Ensemble}, \texttt{Mega Ensemble}, \texttt{RandNet}~\cite{DBLP:conf/sdm/ChenSAT17}); deep learning approaches (e.g., \texttt{RDA}~\cite{DBLP:conf/kdd/ZhouP17}, \texttt{DAGMM}~\cite{DBLP:conf/iclr/ZongSMCLCC18}, \texttt{DeepSVDD}~\cite{DBLP:conf/icml/RuffGDSVBMK18}, and recent deep detectors \texttt{ROBOD}~\cite{DBLP:conf/nips/DingZA22}, \texttt{LUNAR}~\cite{DBLP:conf/aaai/GoodgeHNN22}, \texttt{DTE-C}~\cite{DBLP:conf/iclr/LivernocheJHR24}, and \texttt{TCCM}~\cite{DBLP:journals/corr/abs-2510-18328} evaluated under the same unsupervised protocol); and meta-learning or ensemble-based selectors, including \texttt{MetaOD}~\cite{DBLP:journals/corr/abs-2009-10606}, \texttt{LSCP}~\cite{DBLP:conf/sdm/ZhaoNHL19}, and the state-of-the-art \texttt{ELECT}~\cite{DBLP:conf/icdm/0016ZA22}. We further consider ensemble variants built on \texttt{ELECT} (top-$k$ aggregation and random expansion) to assess whether simple rank-based or hybrid strategies suffice. Finally, we report a supervised greedy oracle, which iteratively selects models using ground-truth labels as an upper bound. Detailed descriptions are provided in App.~\ref{app:subsec:baselines}.


\paragraph{Hyperparameter \& Implementation Details.}
No method uses target labels for tuning. Baselines use fixed defaults (PyOD/Scikit-learn) or the best fixed ensemble size by benchmark AP, while \framework tunes only on labeled meta-training data via leave-one-dataset-out validation. Detailed hyperparameter settings are provided in App.~\ref{app:subsec:hyperparams}, and implementation details are included in App.~\ref{app:subsec:implementation}.


\paragraph{Metrics.} We evaluate all methods using five metrics: Average Precision (AP), Average Rank (AR), ROC-AUC, Precision@$\pi$, and Max F1-score. Details of metrics are provided in App.~\ref{app:metrics}.

\subsection{Main Results}
\label{subsec:main_results}

\begin{table*}[t]
\caption{Main performance comparison on the 39 benchmark datasets. We report Average Precision (AP), Average Rank over datasets (lower is better), ROC-AUC, Precision@$\pi$ where $\pi$ is the number of true anomalies, Max-F1 over all thresholds, and the average selected ensemble size. Best results are in \textbf{bold}; second-best are \underline{underlined}. Method abbreviations: \texttt{Singleton} is random single-model selection, \texttt{Mega Ensemble} averages all 297 detectors, and \texttt{ELECT} Top-$k$ aggregates the top-$k$ models selected by \texttt{ELECT}.}
\label{tab:baselines}
\centering
\resizebox{\linewidth}{!}{%
\begin{tabular}{@{}lllllll@{}}
\toprule
\textbf{Method} & \textbf{AP} $\uparrow$ & \textbf{Rank} $\downarrow$ & \textbf{ROC-AUC} $\uparrow$ & \textbf{Prec@$\pi$} $\uparrow$ & \textbf{Max-F1} $\uparrow$ & \textbf{Ens Size} \\
\midrule
\multicolumn{7}{@{}l}{\textit{Theoretical Upper Bound}} \\
Greedy Oracle & \textcolor{violet}{0.6877} \phantom{$\pm$ 0.0000} & \textcolor{violet}{1.0} \phantom{$\pm$ 0.0000} & \textcolor{violet}{0.8968} \phantom{$\pm$ 0.0000} & \textcolor{violet}{0.6504} \phantom{$\pm$ 0.0000} & \textcolor{violet}{0.6906} \phantom{$\pm$ 0.0000} & 10 \\
\midrule
\multicolumn{7}{@{}l}{\textit{Single Model Baselines}} \\
\texttt{Singleton} & 0.3495 $\pm$ 0.0255 & 147.9 $\pm$ 22.0578 & 0.7081 $\pm$ 0.0248 & 0.3304 $\pm$ 0.0283 & 0.4075 $\pm$ 0.0190 & 1 \\
\texttt{LOF} & 0.3513 $\pm$ 0.0038 & 120.1 $\pm$ 2.2336 & 0.7439 $\pm$ 0.0063 & 0.3297 $\pm$ 0.0037 & 0.4169 $\pm$ 0.0051 & 1 \\
\texttt{Global Best} & 0.3787 $\pm$ 0.0074 & 122.7 $\pm$ 9.0314 & 0.7583 $\pm$ 0.0049 & 0.3593 $\pm$ 0.0093 & 0.4266 $\pm$ 0.0075 & 1 \\
\midrule
\multicolumn{7}{@{}l}{\textit{Na\"ive \& Random Ensembles}} \\
\texttt{IForest} & 0.3858 $\pm$ 0.0016 & 117.0 $\pm$ 5.8119 & 0.7699 $\pm$ 0.0018 & 0.3619 $\pm$ 0.0033 & 0.4337 $\pm$ 0.0022 & 200 \\
\texttt{RandNet} & 0.3460 $\pm$ 0.0018 & 170.6 $\pm$ 2.9136 & 0.6865 $\pm$ 0.0029 & 0.3189 $\pm$ 0.0034 & 0.3927 $\pm$ 0.0027 & 20 \\
\texttt{Mega Ensemble} & 0.3970 $\pm$ 0.0000 & 100.0 $\pm$ 0.0000 & 0.7737 $\pm$ 0.0000 & 0.3782 $\pm$ 0.0000 & 0.4443 $\pm$ 0.0000 & 297 \\
\texttt{Random Ensemble} & 0.3759 $\pm$ 0.0175 & 124.3 $\pm$ 13.6874 & 0.7477 $\pm$ 0.0106 & 0.3608 $\pm$ 0.0172 & 0.4290 $\pm$ 0.0140 & 3 \\
\midrule
\multicolumn{7}{@{}l}{\textit{Deep Learning Baselines}} \\
\texttt{RDA} & 0.2742 $\pm$ 0.0065 & 211.9 $\pm$ 11.4741 & 0.7063 $\pm$ 0.0064 & 0.2721 $\pm$ 0.0090 & 0.3515 $\pm$ 0.0068 & 1 \\
\texttt{DAGMM} & 0.2958 $\pm$ 0.0124 & 221.6 $\pm$ 6.8508 & 0.6676 $\pm$ 0.0134 & 0.3013 $\pm$ 0.0135 & 0.3681 $\pm$ 0.0129 & 1 \\
\texttt{DeepSVDD} & 0.2073 $\pm$ 0.0115 & 247.5 $\pm$ 6.8516 & 0.5905 $\pm$ 0.0164 & 0.2116 $\pm$ 0.0147 & 0.2968 $\pm$ 0.0073 & 1 \\
\texttt{ROBOD} & 0.3135 $\pm$ 0.0026 & 208.0 $\pm$ 2.1602 & 0.6665 $\pm$ 0.0001 & 0.2973 $\pm$ 0.0024 & 0.3621 $\pm$ 0.0002 & 1 \\
\texttt{LUNAR} & 0.3024 $\pm$ 0.0045 & 172.0 $\pm$ 18.6815 & 0.6566 $\pm$ 0.0062 & 0.3026 $\pm$ 0.0098 & 0.3745 $\pm$ 0.0074 & 1 \\
\texttt{DTE-C} & 0.3144 $\pm$ 0.0012 & 199.7 $\pm$ 9.2376 & 0.7469 $\pm$ 0.0087 & 0.3083 $\pm$ 0.0033 & 0.3971 $\pm$ 0.0016 & 1 \\
\texttt{TCCM} & 0.2929 $\pm$ 0.0144 & 188.3 $\pm$ 28.2902 & 0.6745 $\pm$ 0.0126 & 0.2834 $\pm$ 0.0226 & 0.3690 $\pm$ 0.0111 & 1 \\
\midrule
\multicolumn{7}{@{}l}{\textit{Meta-Learning \& Ensemble Methods}} \\
\texttt{LSCP} & 0.3484 $\pm$ 0.0173 & 124.3 $\pm$ 11.1161 & 0.7560 $\pm$ 0.0096 & 0.3441 $\pm$ 0.0208 & 0.4251 $\pm$ 0.0123 & 1 \\
\texttt{MetaOD} & 0.3989 $\pm$ 0.0024 & 101.0 $\pm$ 7.6594 & 0.7547 $\pm$ 0.0014 & 0.3746 $\pm$ 0.0032 & 0.4392 $\pm$ 0.0022 & 1 \\
\texttt{ELECT}+Random & 0.3981 $\pm$ 0.0060 & 102.2 $\pm$ 9.2232 & 0.7719 $\pm$ 0.0068 & 0.3778 $\pm$ 0.0103 & 0.4449 $\pm$ 0.0067 & 10 \\
\texttt{ELECT} (Top-1) & 0.4069 $\pm$ 0.0063 & 85.8 $\pm$ 7.8712 & 0.7734 $\pm$ 0.0046 & \underline{0.3861} $\pm$ 0.0059 & 0.4519 $\pm$ 0.0051 & 1 \\
\texttt{ELECT} (Top-10) & \underline{0.4117} $\pm$ 0.0050 & \underline{83.2} $\pm$ 6.8118 & \underline{0.7785} $\pm$ 0.0038 & 0.3856 $\pm$ 0.0055 & \underline{0.4546} $\pm$ 0.0043 & 10 \\
\midrule
\framework (\textbf{Ours}) & \textbf{0.4308} $\pm$ 0.0064 & \textbf{59.3} $\pm$ 6.9610 & \textbf{0.7867} $\pm$ 0.0045 & \textbf{0.4042} $\pm$ 0.0063 & \textbf{0.4681} $\pm$ 0.0069 & 2.2 \\
\bottomrule
\end{tabular}
}
\end{table*}

Table~\ref{tab:baselines} compares \framework with unsupervised baselines on 39 benchmark datasets. The per-dataset results are in App.~\ref{app:sec:detailed_experiments}. Overall, \framework achieves the best performance across metrics. Notably, it consistently outperforms the strongest meta-learning baseline \texttt{ELECT}, despite both methods sharing the same primary detector, demonstrating the effectiveness of our sequential partner selection strategy.

Na\"ive aggregation strategies perform substantially worse. Methods that average many detectors (\texttt{Mega Ensemble}) or randomly form ensembles (\texttt{Random Ensemble}) fail to filter weak or redundant models, highlighting the necessity of informed selection. Similarly, single detectors and static meta-learners such as \texttt{Global Best} fall behind adaptive model selection methods.

Deep learning baselines show limited effectiveness on these fully unsupervised tabular anomaly detection tasks, consistent with large-scale tabular studies~\cite{DBLP:conf/nips/GrinsztajnOV22,DBLP:journals/inffus/Shwartz-ZivA22} and ADBench~\cite{DBLP:conf/nips/HanHHJ022}. This should not be read as a general limitation of deep anomaly detection; rather, it reflects the difficulty of tuning neural objectives without labels on heterogeneous tabular datasets with small sample sizes, mixed feature types, and severe class imbalance. Recent methods such as \texttt{ROBOD}, \texttt{LUNAR}, \texttt{DTE-C}, and \texttt{TCCM} improve over older deep baselines but still trail \framework, suggesting that cross-family model selection is more effective here than committing to a single neural architecture family.

We also compare with ensemble variants built on \texttt{ELECT}. Simply aggregating top-ranked models provides only marginal improvements over the single-model selector, whereas \framework's context-aware expansion yields clear gains. Across additional evaluation metrics, \framework remains among the top performers while using compact, adaptive ensembles, demonstrating both effectiveness and efficiency.

\subsection{Ablation Study}
\label{subsec:ablation}


\begin{table}[t]
\fontsize{8}{8}\selectfont
\caption{Ablation study on the 39 benchmark datasets. $\Delta$AP measures the degradation relative to the full model. Setting $\beta=0$ removes the redundancy/diversity discount, setting $\lambda_{\mathrm{fam}}=0$ removes family-risk regularization, and the single-part gain model replaces the two-part classifier--regressor architecture with one direct gain predictor.}
\label{tab:ablation}
\centering
\setlength{\tabcolsep}{4pt}
\begin{tabular}{lccc}
\toprule
\textbf{Variant} & \textbf{AP} $\uparrow$ & \textbf{Rank} $\downarrow$ & $\Delta$\textbf{AP} \\
\midrule
w/o diversity ($\beta{=}0$) & 0.4185 & 77 & $-0.0169$ \\
w/o family-risk ($\lambda_{\text{fam}}{=}0$) & 0.3995 & 72 & $-0.0359$ \\
single-part gain model & 0.4133 & 87 & $-0.0221$ \\
\midrule
\framework & \textbf{0.4308} & \textbf{59.3} & — \\
\bottomrule
\end{tabular}
\end{table}

To understand the contribution of each component in \framework, we perform ablations by selectively disabling key mechanisms. Table~\ref{tab:ablation} shows the ablation study. The full model performs best overall.

Removing $\beta$ shows a performance decrease, as $\beta$ controls intra-family similarity while $\lambda_{\text{fam}}$ controls family-level risk. Removing family-risk regularization ($\lambda_{\text{fam}}=0$) causes the largest performance drop, showing that family-aware risk control is the most critical component for robust ensemble construction. Replacing the two-part meta-model with a single gain predictor leads to a consistent but moderate degradation, supporting our design choice of separating improvement probability and magnitude.

A detailed sensitivity analysis of all proxy objective hyperparameters ($\beta$, $\lambda_{\text{fam}}$, $\tau_1$, $\tau_2$) across search grids and 3 random seeds is provided in App.~\ref{app:subsec:hyperparams}. Additional family-granularity and single-family analyses are in App.~\ref{app:subsec:family_ablation} and App.~\ref{app:sec:pool_size_analysis}.

\begin{figure*}[t]
\centering
\includegraphics[width=0.8\linewidth]{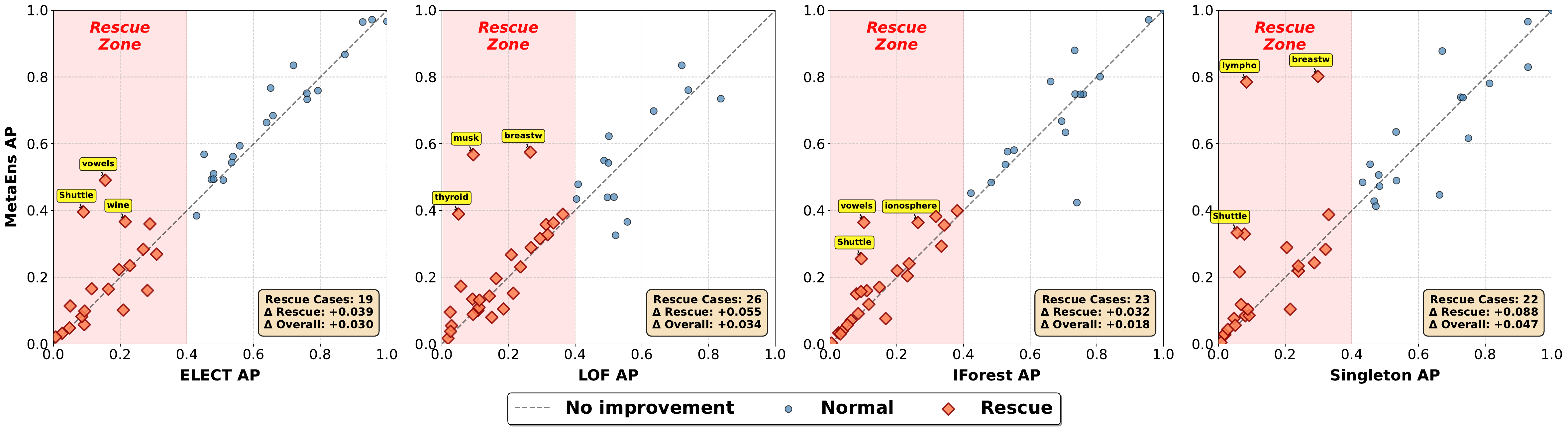}
\caption{Robustness analysis across four different primary selectors: \texttt{ELECT}, \texttt{LOF}, \texttt{IForest}, and \texttt{Random Selection}. Each panel compares the primary model's performance ($x$-axis) against the final \framework ensemble ($y$-axis). Points above the diagonal indicate improvement. The shaded red ``Rescue Zone'' highlights where the primary model fails (AP $< 0.4$). \framework consistently rescues performance in these failure modes across all primary selectors, demonstrating that its diverse partner selection logic is robust and selector-agnostic.}
\label{fig:resilience_scatter_grid}
\end{figure*}

\subsection{Statistical Significance}
\label{subsec:statistical_significance}
To validate the performance improvements of \framework, we conduct paired one-sided Wilcoxon signed-rank tests comparing our method against all baselines across the 39 datasets. \framework achieves statistically significant AP improvements over all baselines, including recent deep detectors. The rank-based view is important because the benchmark is heterogeneous: average AP can be dominated by easier datasets with high anomaly prevalence, whereas rank summarizes whether a method is consistently competitive across datasets with different contamination rates and feature types. Full $p$-values and win rates are provided in App.~\ref{app:sec:statistical_significance}.

\subsection{Modality Transfer to Image and Text Anomaly Detection}
\label{subsec:modality_transfer}
Because \framework uses only normalized score vectors, it transfers beyond tabular data. On 20 ADBench image/text extracted-feature datasets (15 MVTec-AD, 5 text), \framework improves over the strongest baseline overall and on images ($+0.0257$ AP), with significant wins across baselines; see App.~\ref{app:subsec:modality_transfer}. This supports the score-only design: the meta-model does not depend on raw feature dimensionality, input modality, or detector internals, and can reuse the same state representation for image, text, and tabular anomaly scores.

\subsection{Robustness to Initialization}
\label{sec:robustness}

A central question is whether \framework's gains depend on a specific primary selector or on strong initial models. To investigate this, we evaluate \framework under diverse initialization strategies and analyze its behavior when the starting model performs poorly.

We pair \framework with heterogeneous primary selectors, including the meta-learning method \texttt{ELECT}, density-based method \texttt{LOF}, tree-based \texttt{IForest}, and a random singleton baseline. For each configuration, we compare the primary selector's AP with the final AP achieved by \framework. Detailed scatter plots and per-selector analyses are shown in Fig.~\ref{fig:resilience_scatter_grid}.

Across all selectors, \framework consistently improves over the starting model, demonstrating that its partner selection mechanism is selector-agnostic and not tied to a particular initialization strategy. Improvements are especially pronounced in challenging cases where the primary model underperforms. Rather than propagating initial errors, \framework effectively recovers performance by selecting complementary detectors from diverse algorithmic families. Even when initialized with weak or biased selectors, the method constructs strong ensembles, indicating that gains stem from diversity-aware partner selection rather than reliance on the primary model quality.


\subsection{Failure Mode Analysis}
\label{subsec:failure_mode}

While \framework consistently outperforms baselines on average, per-dataset results (App.~\ref{app:sec:detailed_experiments}) reveal systematic underperformance on a small number of datasets. Notably, \framework trails strong single detectors on low-dimensional datasets such as \texttt{Glass}, \texttt{HeartDisease}, \texttt{Cardiotocography}, and \texttt{Vertebral} (5--13 features), where classical detectors with strong inductive biases (e.g., \texttt{LOF}, \texttt{IForest}) are already near-optimal, leaving little margin for ensemble improvement. This is a known limitation of ensemble methods in general~\cite{DBLP:journals/sigkdd/ZimekCS13}.

Figure~\ref{fig:resilience_scatter_grid} further identifies failure modes via the \emph{Rescue Zone} (primary AP $< 0.4$): when all candidate detectors in the pool perform poorly, no complementary partner exists and \framework cannot recover. In such cases, adaptive early stopping prevents further degradation. Distribution shift between test and meta-training datasets is partially mitigated by our score-based, distribution-agnostic state representation (Table~\ref{app:tab:features}): the meta-model learns patterns of inter-detector behavior from normalized score vectors rather than raw input features, enabling transfer across heterogeneous regimes.

\subsection{Model Diversity Analysis}
\label{subsec:model_diversity}

To illustrate the difference between \framework and baseline selection strategies, we project the 297-model candidate pool into a two-dimensional space using \texttt{t-SNE}~\cite{DBLP:journals/JMLR/MaatenH08} based on prediction-correlation distance. This embedding reveals that models naturally cluster by algorithmic family, as detectors within the same family tend to produce similar outlier rankings.

\begin{figure*}[t]
\centering
\includegraphics[width=0.8\linewidth]{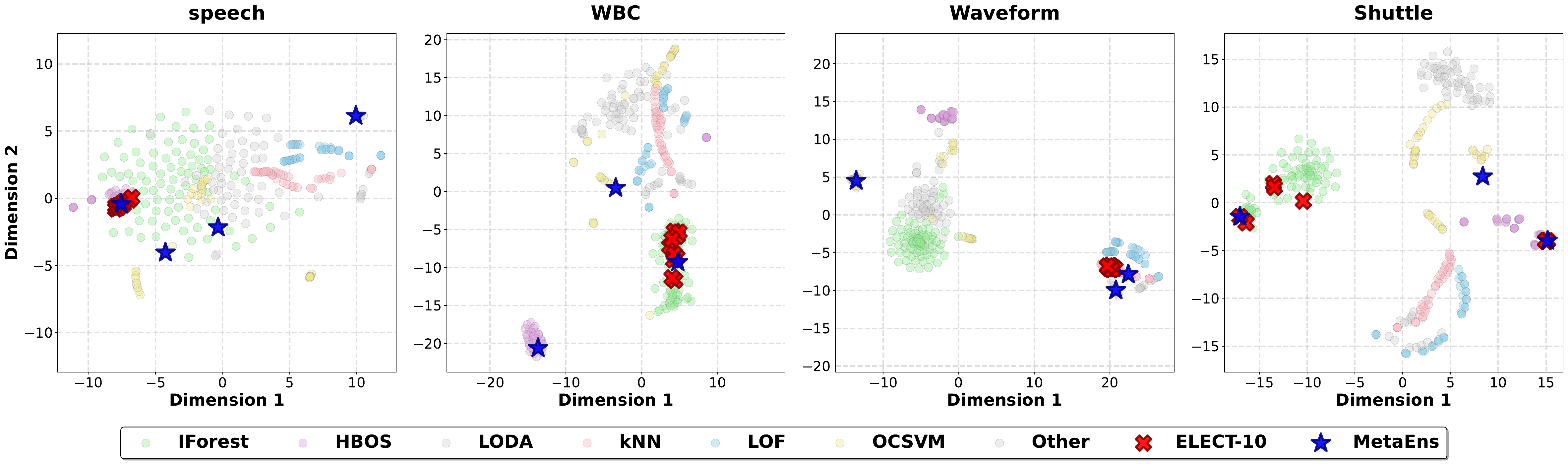}
\caption{Model diversity visualization using \texttt{t-SNE} projection across four datasets. \texttt{ELECT-10} selections tend to cluster within a single family, whereas \framework selects models spanning multiple families, indicating greater ensemble diversity.}
\label{fig:diversity_tsne}
\end{figure*}

Figure~\ref{fig:diversity_tsne} shows the resulting visualization on four representative datasets (\textbf{Speech}, \textbf{WBC}, \textbf{Waveform}, and \textbf{Shuttle}), overlaid with the models selected in our experiments. Background points correspond to all 297 detectors, colored by family, and exhibit clear family-level clustering. In all cases, \texttt{ELECT-10} selects models concentrated within a single family cluster (e.g., \texttt{HBOS} for \textbf{Speech} and \texttt{IForest} for \textbf{WBC}), indicating limited diversity. In contrast, \framework selects models distributed across multiple distinct family clusters. For example, on \textbf{Speech}, selections span four families (\texttt{HBOS}, \texttt{IForest}, \texttt{OCSVM}, \texttt{ABOD}), while on \textbf{WBC}, selections cover three (\texttt{IForest}, \texttt{OCSVM}, \texttt{HBOS}).

This visualization supports the observation that \framework does not rely solely on the initial model chosen by \texttt{ELECT}. Although both methods share the same primary detector, \framework's family-risk regularization encourages selection of complementary models from different algorithm families. The dispersion of selected detectors suggests that the framework captures diverse decision patterns, which is particularly important in unsupervised settings where validation labels are unavailable. Quantitative diversity measurements and additional discussion are provided in App.~\ref{app:sec:mode_diversity}.

\subsection{Pool Size Analysis}
\label{subsec:pool_size}
A detailed analysis of the effect of candidate pool size is provided in App.~\ref{app:sec:pool_size_analysis}. The study shows how \framework's performance evolves as the number of available detectors increases, demonstrating that the method scales well and remains stable while benefiting from greater model diversity.

\subsection{Effects of Meta-model Architectures}
\label{subsec:meta_model_analysis}
We study the choice of meta-model architectures in App.~\ref{app:subsec:meta_model_analysis}. The comparison evaluates several alternative learning families for marginal gain prediction and candidate filtering, showing that the default \texttt{ExtraTrees} configuration achieves the best balance between predictive performance and computational efficiency.

\subsection{Effect of Expanded Model Pools}
\label{subsec:expanded_model_pool}
To evaluate \framework's scalability and robustness to model pool composition, we conduct additional experiments with an expanded pool of 310 models that incorporates neural networks. A detailed analysis of the expanded model pool is provided in App.~\ref{app:subsec:expanded_model_pool}.

\subsection{Effect of Ensemble Size}
\label{subsec:ensemble_size}
To understand how ensemble size affects performance, we analyze the relationship between the number of models and detection quality. A detailed analysis of the effect of ensemble size is provided in App.~\ref{app:subsec:ensemble_size}.
\section{Related Work}
\label{sec:relatedwork}

\subsection{Unsupervised Outlier Detection}
Unsupervised outlier detection has been studied across diverse methodological paradigms. Statistical methods such as \texttt{HBOS}~\cite{DBLP:conf/ijcnn/AryalBRS21} identify anomalies in low-density regions under feature independence assumptions. Density- and distribution-based methods, including \texttt{LOF}~\cite{DBLP:conf/sigmod/BreunigKNS00}, \texttt{COPOD}~\cite{DBLP:conf/icdm/LiZBIH20}, and \texttt{ECOD}~\cite{DBLP:journals/tkde/LiZHBIC23}, quantify deviations using local density or tail probabilities of empirical distributions. Tree-based methods such as \texttt{IForest}~\cite{DBLP:conf/icdm/LiuTZ08} exploit random partitioning to isolate anomalies, while distance-based methods including \texttt{kNN}~\cite{DBLP:conf/sdm/Chehreghani16} and \texttt{ODIN}~\cite{DBLP:conf/icpr/HautamakiKF04} detect outliers via spatial isolation. One-class classifiers, such as \texttt{OCSVM}~\cite{DBLP:conf/nips/ScholkopfWSSP99} and \texttt{SVDD}~\cite{DBLP:journals/ml/TaxD04}, learn decision boundaries enclosing normal data, whereas subspace-based methods like \texttt{PCA}~\cite{DBLP:conf/pkdd/ChapelF14} rely on reconstruction errors.
More recently, deep learning–based methods learn expressive representations of normality. These include autoencoders (\texttt{AE}s)~\cite{DBLP:conf/ijcai/GoodgeHNN20}, variational autoencoders (\texttt{VAE}s)~\cite{DBLP:conf/www/XuCZLBLLZPFCWQ18}, and generative adversarial networks (\texttt{GAN}s)~\cite{DBLP:conf/icdm/LimLTCRE18}. Extensions such as \texttt{Deep SVDD}~\cite{DBLP:conf/icml/RuffGDSVBMK18}, \texttt{DAGMM}~\cite{DBLP:conf/iclr/ZongSMCLCC18}, and \texttt{RDA}~\cite{DBLP:conf/kdd/ZhouP17} integrate representation learning with one-class objectives or density estimation to improve robustness. Recent methods further explore graph neural representations (\texttt{LUNAR})~\cite{DBLP:conf/aaai/GoodgeHNN22}, diffusion-based anomaly scoring (\texttt{DTE-C})~\cite{DBLP:conf/iclr/LivernocheJHR24}, and flow-matching formulations (\texttt{TCCM})~\cite{DBLP:journals/corr/abs-2510-18328}. Transformer-based models~\cite{DBLP:conf/cikm/KimLH24} have also been employed to capture complex dependencies. Despite their effectiveness, these methods are often sensitive to architectural choices and hyperparameters, motivating ensemble-based strategies to improve generalization.

\subsection{Ensembles for Outlier Detection}
Ensemble methods aim to improve robustness and stability by combining multiple detectors~\cite{DBLP:journals/sigkdd/AggarwalS15}. Common aggregation strategies include score averaging, ranking, and voting; however, naively combining all detectors often incurs high computational cost, performance degradation due to weak models, and limited adaptivity. Several ensemble designs address these issues implicitly. Feature-bagging ensembles promote diversity through random subspaces~\cite{DBLP:conf/icdm/NotoBS10,DBLP:conf/kdd/LazarevicK05}, while tree-based ensembles such as \texttt{IForest}~\cite{DBLP:conf/icdm/LiuTZ08} embed diversity via randomized construction. Sequential ensembles refine performance by reweighting instances across rounds, as exemplified by \texttt{XGBOD}~\cite{DBLP:conf/ijcnn/ZhaoH18}. Stacking-based approaches combine heterogeneous detectors, including autoencoder~\cite{DBLP:conf/sdm/ChenSAT17} and \texttt{GAN} ensembles~\cite{DBLP:conf/aaai/HanCL21}, to preserve complementary behaviors. ROBOD~\cite{DBLP:conf/nips/DingZA22} specifically addresses hyperparameter sensitivity in deep outlier detection by constructing a scalable homogeneous hyper-ensemble over neural detectors. Yet, most existing methods rely on fixed aggregation schemes, a single architecture family, or require constructing the full ensemble upfront, limiting adaptivity in unsupervised settings. Unlike these approaches, \framework focuses on label-free model selection over heterogeneous detector families rather than score reweighting within a fixed architecture.

\subsection{Model Selection}
Model selection is well established in supervised learning via information-theoretic criteria such as AIC and BIC~\cite{DBLP:journals/annstat/Schwarz78}, cross-validation~\cite{DBLP:journals/jrstatsocb/Stone74}, structural risk minimization~\cite{DBLP:books/daglib/0097035}, as well as modern hyperparameter optimization techniques including Bayesian optimization~\cite{DBLP:conf/nips/SnoekLA12} and automated machine learning frameworks~\cite{DBLP:conf/nips/FeurerKESBH15}. In contrast, unsupervised outlier model selection is considerably more challenging due to the lack of labels, extreme class imbalance, and heterogeneous anomaly patterns. Early approaches rely on internal validation heuristics derived from score distributions or stability under perturbations~\cite{DBLP:journals/corr/Goix16,DBLP:journals/tkdd/MarquesCSZ20}, which often correlate weakly with true detection performance.
Recent studies explore meta-learning approaches~\cite{DBLP:journals/air/VilaltaD02,DBLP:journals/pami/HospedalesAMS22} that leverage labeled meta-datasets to generalize model selection across tasks. Representative methods include \texttt{MetaOD}~\cite{DBLP:journals/corr/abs-2009-10606} and \texttt{ELECT}~\cite{DBLP:conf/icdm/0016ZA22}, which focus primarily on selecting individual detectors. Ensemble-based model selection remains underexplored: \texttt{LSCP}~\cite{DBLP:conf/sdm/ZhaoNHL19} performs instance-wise detector selection but assumes local data consistency and requires constructing the full ensemble in advance. In contrast, \framework performs sequential ensemble model selection without assuming local consistency and supports adaptive early stopping, enabling compact, high-performing ensembles in fully unsupervised settings.

\section{Conclusion}
\label{sec:conclusion}
We address the problem of unsupervised ensemble outlier model selection, where ensembles must be constructed without access to labels. We propose \framework, a meta-learned framework that predicts marginal ensemble gains and combines these with a submodular-inspired proxy objective to guide adaptive ensemble construction with early stopping. Extensive experiments on 39 real-world datasets show that \framework can consistently outperform strong unsupervised baselines while selecting substantially smaller ensembles, highlighting the importance of family-level diversification and risk control in unsupervised settings. 

Future work includes exploring richer meta-representations to improve gain prediction under distribution shift, extending the framework to streaming or non-stationary data, and developing theoretical or uncertainty-aware variants of the proxy objective to better understand unsupervised ensemble selection.

\section*{Broader Impact}

Unsupervised outlier detection is used widely in high-impact domains such as fraud detection, cybersecurity, healthcare monitoring, and scientific data analysis. By enabling adaptive and data-driven ensemble construction without requiring labeled data, \framework has the potential to improve the robustness and reliability of outlier detection systems deployed in practice. In particular, the ability to construct compact ensembles can reduce computational cost and energy consumption, which is beneficial for large-scale or resource-constrained settings.
\section*{Limitations}
This work has several limitations. First, \framework relies on labeled meta-datasets to learn transferable patterns of marginal ensemble gain. While our experiments suggest strong generalization across diverse datasets, performance may degrade if test tasks differ substantially from the
meta-training distribution; empirically, underperformance is most pronounced
on low-dimensional datasets ($d \leq 13$) where single detectors are already
near-optimal (see Sec.~\ref{subsec:failure_mode}). The score-based,
distribution-agnostic state representation partially mitigates this risk, as
confirmed by statistically significant improvements across heterogeneous
benchmarks spanning contamination $0.20\%$--$45.78\%$ and dimensionality
$5$--$1{,}555$. Second, the family-risk regularization depends on a predefined mapping of detectors to algorithmic families, which may be coarse and require domain knowledge to define for new detectors. Finally, we focus on batch settings and do not address streaming or non-stationary data, which we leave for future work.
\bibliographystyle{icml2026}
\bibliography{ref}
\counterwithin{figure}{section}
\counterwithin{table}{section}
\appendix
\section{Appendix}

\subsection{Ablation on Family Definition}
\label{app:subsec:family_ablation}

To assess the sensitivity of \framework to the granularity of the family definition, we compare three variants: (1)~no family-risk penalty ($\lambda_{\text{fam}}=0$), (2)~a \emph{coarse-grained} definition mapping the 8 base algorithms into 3 broad super-families, and (3)~our default \emph{fine-grained} definition using all 8 base families independently.

The coarse-grained grouping maps algorithms as follows:
\textit{Density/Proximity} (\texttt{LOF}, \texttt{kNN}, \texttt{COF}),
\textit{Isolation/Tree} (\texttt{IForest}), and
\textit{Linear/Probabilistic} (\texttt{HBOS}, \texttt{OCSVM}, \texttt{LODA},
\texttt{ABOD}).

\begin{table}[h]
\centering
\footnotesize
\caption{Ablation on family definition granularity.
$\Delta$AP measures performance degradation relative to the full model.}
\label{tab:family_ablation}
\begin{tabular}{@{}lccc@{}}
\toprule
\textbf{Variant} & \textbf{AP} $\uparrow$ & \textbf{Rank} $\downarrow$ & $\Delta$\textbf{AP} \\
\midrule
w/o family-risk & 0.3995 & 72.0 & $-$0.0313 \\
Coarse-grained & 0.4196 & 65.0 & $-$0.0112 \\
\framework\ Fine-grained & \textbf{0.4308} & \textbf{59.3} & --- \\
\bottomrule
\end{tabular}
\end{table}

Fine-grained family definitions yield the strongest performance.
Even the coarse-grained variant substantially outperforms the no-penalty baseline ($\Delta$AP $= -0.0112$ vs.\ $-0.0313$), demonstrating that \framework does not require a perfectly optimal taxonomy. However, broad groupings can cause \emph{collateral penalties}: for instance, a strong \texttt{kNN} candidate may be suppressed because a \texttt{LOF} model underperformed, despite their differing sensitivity to local versus global anomaly patterns. The fine-grained definition ensures that a family's historical failure penalizes only its own hyperparameter variants, without unfairly suppressing distinct but conceptually related algorithms.

\subsection{Hyperparameter Settings}
\label{app:subsec:hyperparams}

\paragraph{Baseline tuning protocol.}
To preserve the fully unsupervised evaluation setting, target-dataset labels are never used to tune any baseline. Single-model baselines are evaluated with fixed default configurations from standard implementations (e.g., PyOD and Scikit-learn) across all datasets. For baselines with an ensemble-size parameter $k$, we report the best fixed $k$ according to average AP over the benchmark, giving these baselines an oracle advantage over normal unsupervised deployment. The supervised greedy oracle is reported only as an upper bound and is not a deployable unsupervised method.

\paragraph{\framework hyperparameters.}
We configure the following hyperparameters, which remain fixed across all 39 test datasets to ensure fair comparison. For primary model selection via \texttt{ELECT}~\cite{DBLP:conf/icdm/0016ZA22}, we adopt the default configuration. For marginal gain prediction, we train a two-part meta-model using \texttt{ExtraTrees}: (i) a classifier component for predicting improvement probability ($y_p = \mathbb{I}(g > 0)$) with balanced class weights, and (ii) a regression component for predicting improvement magnitude ($y_m = \max(0, g)$) optimized for MAE loss. For sequential selection, we set acceptance thresholds $\tau_1$ and $\tau_2$ for first-partner selection and iterative expansion, respectively. For diversity modulation, we configure discount strength $\beta$, family-risk weight $\lambda_{\text{fam}}$, and Jaccard similarity top-$k$ threshold $k_{\text{top}}$. The ensemble size budget is $\eta$ with adaptive early stopping. All hyperparameters are tuned via leave-one-out cross-validation on meta-training data. Table~\ref{tab:hp_grid} lists the search grids for all tunable proxy objective hyperparameters.

\begin{table*}[t]
\centering
\small
\caption{Hyperparameter search grids.}
\label{tab:hp_grid}
\begin{tabular}{@{}llp{8cm}@{}}
\toprule
\textbf{Hyperparameter} & \textbf{Role} & \textbf{Search Grid} \\
\midrule
$\tau_1, \tau_2$ & Acceptance thresholds
    & $\{-0.01, 0, 0.001/0.005, 0.01, 0.05, 0.1\}$ \\
$\beta$ & Diversity discount strength
    & $\{0, 0.5, 1, 2, 3, 5, 10\}$ \\
$\lambda_{\text{fam}}$ & Family-risk weight
    & $\{0, 0.2, 0.4, 0.6, 0.8, 1.0, 1.5\}$ \\
\bottomrule
\end{tabular}
\end{table*}

\paragraph{Sensitivity analysis.}
Table~\ref{tab:hp_grid} summarizes the search grids used.  ables~\ref{tab:sens_beta}, \ref{tab:sens_lfam}, \ref{tab:sens_risk_percentile}, and \ref{tab:sens_tau} report sensitivity across 39 datasets with 3 seeds. $\beta$ is effectively insensitive across the entire search range (AP variance ${<}0.0005$). $\lambda_{\text{fam}}$ must be at least $0.2$ to activate family-risk regularization; once active, all larger values yield stable performance, consistent with Table~\ref{tab:ablation}. Setting $\tau$ too high causes premature early stopping; the default small positive threshold ($0.001/0.005$) does not require fine-tuning. For the family-risk percentile, the 10th percentile is a fixed global lower-tail choice rather than a target-dataset-tuned parameter. It penalizes families with occasional catastrophic negative gains while preserving families whose average behavior may still contribute orthogonal diversity.

\begin{table}[h]
\centering
\footnotesize
\setlength{\tabcolsep}{4pt}
\caption{Sensitivity to $\beta$ (diversity discount strength).}
\label{tab:sens_beta}
\begin{tabular}{@{}lccc@{}}
\toprule
$\beta$ & AP (Mean $\pm$ Std) $\uparrow$ & ROC-AUC $\uparrow$ & Rank $\downarrow$ \\
\midrule
0           & $0.4303 \pm 0.0057$ & 0.7843 & 60.5 \\
0.5         & $0.4304 \pm 0.0057$ & 0.7842 & 60.2 \\
1           & $0.4304 \pm 0.0057$ & 0.7842 & 60.2 \\
2           & $0.4308 \pm 0.0064$ & 0.7867 & 59.3 \\
3~(def.)    & $\mathbf{0.4308 \pm 0.0064}$ & \textbf{0.7867} & \textbf{59.3} \\
5           & $0.4306 \pm 0.0058$ & 0.7841 & 59.8 \\
10          & $0.4308 \pm 0.0058$ & 0.7841 & 59.8 \\
\bottomrule
\end{tabular}
\end{table}

\begin{table}[h]
\centering
\footnotesize
\setlength{\tabcolsep}{4pt}
\caption{Sensitivity to the family-risk percentile $\tau_{\text{risk}}$ used in
Eq.~\ref{eq:family-risk}. The default $10\%$ threshold matches the main result reported in Table~\ref{tab:baselines}.}
\label{tab:sens_risk_percentile}
\begin{tabular}{@{}lcc@{}}
\toprule
$\tau_{\text{risk}}$ & AP (Mean $\pm$ Std) $\uparrow$ & Median Rank $\downarrow$ \\
\midrule
$0\%$ (no risk penalty)      & $0.4269 \pm 0.0066$ & 67.7 \\
$5\%$                         & $0.4272 \pm 0.0066$ & 67.7 \\
$10\%$~(def.)                 & $\mathbf{0.4308 \pm 0.0064}$ & \textbf{59.3} \\
$25\%$                        & $0.3368 \pm 0.0048$ & 143.3 \\
$50\%$ (median)               & $0.3142 \pm 0.0087$ & 170.0 \\
\bottomrule
\end{tabular}
\end{table}

Table~\ref{tab:sens_risk_percentile} shows that \framework is robust in the lower-tail regime from $0\%$ to $10\%$, indicating that performance does not depend on a precisely fine-tuned percentile. In contrast, aggressive thresholds such as $25\%$ and $50\%$ substantially degrade AP and rank because they over-penalize candidate families and suppress useful ensemble diversity. The 10th percentile, therefore, provides a conservative trade-off between mitigating downside risk and preserving cross-family complementarity.

\begin{table}[h]
\centering
\footnotesize
\setlength{\tabcolsep}{4pt}
\caption{Sensitivity to $\lambda_{\text{fam}}$ (family-risk weight).}
\label{tab:sens_lfam}
\begin{tabular}{@{}lccc@{}}
\toprule
$\lambda_{\text{fam}}$ & AP (Mean $\pm$ Std) $\uparrow$ & ROC-AUC $\uparrow$ & Rank $\downarrow$ \\
\midrule
0              & $0.3995 \pm 0.0041$ & 0.7111 & 72.0 \\
0.2~(def.)     & $\mathbf{0.4308 \pm 0.0064}$ & \textbf{0.7867} & \textbf{59.3} \\
0.4            & $0.4231 \pm 0.0065$ & 0.7802 & 68.3 \\
0.6            & $0.4231 \pm 0.0065$ & 0.7802 & 68.3 \\
0.8            & $0.4231 \pm 0.0065$ & 0.7802 & 68.3 \\
1.0            & $0.4231 \pm 0.0065$ & 0.7802 & 68.3 \\
1.5            & $0.4231 \pm 0.0065$ & 0.7802 & 68.3 \\
\bottomrule
\end{tabular}
\end{table}

\begin{table}[h]
\centering
\footnotesize
\setlength{\tabcolsep}{3pt}
\caption{Sensitivity to $\tau_1, \tau_2$ (acceptance thresholds).}
\label{tab:sens_tau}
\begin{tabular}{@{}lccc@{}}
\toprule
$\tau_1,\ \tau_2$ & AP (Mean $\pm$ Std) $\uparrow$ & ROC-AUC $\uparrow$ & Rank $\downarrow$ \\
\midrule
$-0.01,\ {-}0.01$         & $0.4257 \pm 0.0060$ & 0.7836 & 68.7 \\
$0,\ 0$                   & $0.4257 \pm 0.0060$ & 0.7836 & 68.7 \\
$0.001,\ 0.005$~(def.)    & $\mathbf{0.4308 \pm 0.0064}$ & \textbf{0.7867} & \textbf{59.3} \\
$0.01,\ 0.01$             & $0.4041 \pm 0.0052$ & 0.7461 & 86.8 \\
$0.05,\ 0.05$             & $0.4133 \pm 0.0000$ & 0.7792 & 79.1 \\
$0.1,\ 0.1$               & $0.4133 \pm 0.0000$ & 0.7792 & 79.1 \\
\bottomrule
\end{tabular}
\end{table}

\subsection{Ablation on Ensemble Score Aggregation}
\label{app:subsec:score_combiner_ablation}

Equation~\ref{eq:outlier_score_set} defines the ensemble score as the mean of member detector scores. We use the mean because it is the standard and most stable aggregation rule in unsupervised outlier ensembles: it preserves the continuous ranking signal contributed by all selected detectors while reducing dependence on any single base model. In contrast, max aggregation is highly sensitive to isolated false positives from a poorly calibrated detector, while min aggregation requires near-unanimous agreement and is therefore sensitive to false negatives. Median aggregation is more robust to extreme scores, but can discard fine-grained score variation from competent detectors.

To verify this design choice, we conduct a focused aggregation ablation across three random seeds on the 39 benchmark datasets, changing only the operator used to combine selected detector scores. Table~\ref{tab:score_combiner_ablation} shows that mean aggregation achieves the best AP. It significantly outperforms max and min aggregation under a one-sided Wilcoxon test at $\alpha=0.05$. The gap between mean and median is positive but not statistically significant, indicating that the median is a reasonable robust alternative but not a better default in our setting.

\begin{table}[h]
\centering
\footnotesize
\caption{Ablation on ensemble score aggregation. $p$-values are from one-sided Wilcoxon tests against mean aggregation; $^*$ indicates significance at $\alpha=0.05$.}
\label{tab:score_combiner_ablation}
\begin{tabular}{lccc}
\toprule
\textbf{Combiner} & \textbf{AP (Mean $\pm$ Std)} $\uparrow$ & $\Delta$\textbf{AP} & \textbf{$p$-value} \\
\midrule
Mean (default) & \textbf{0.4308}$\pm$0.0064 & --- & --- \\
Median         & 0.4182$\pm$0.0028          & $-$0.0122 & 0.377 \\
Max            & 0.3778$\pm$0.0251          & $-$0.0526 & 0.049$^*$ \\
Min            & 0.3736$\pm$0.0057          & $-$0.0567 & 0.022$^*$ \\
\bottomrule
\end{tabular}
\end{table}

These results support mean aggregation as the most stable default for \framework. This focused ablation isolates the score-combining operator and is therefore complementary to the main benchmark results in Table~\ref{tab:baselines}, where \framework achieves AP $0.4308\pm0.0064$ and average rank $59.3\pm6.96$.

\subsection{Ablation on Training Trajectory Strategies}
\label{app:subsec:trajectory_ablation}

We evaluate two alternative offline training designs against the default oracle greedy rollout: (A)~\emph{$\varepsilon$-greedy extension}, where the greedy argmax is replaced by random sampling with probability $\varepsilon$; and (B)~\emph{reduced meta-dataset coverage}, using 25/50/75\% of labeled meta-datasets. Results are averaged over 3 seeds on the 39-dataset benchmark.

\begin{table}[h]
\centering
\caption{Ablation on training trajectory strategies.}
\label{tab:ablation_trajectory}
\begin{tabular}{lcc}
\toprule
\textbf{Strategy} & \textbf{Mean AP} $\uparrow$ & \textbf{Avg.\ Rank} $\downarrow$ \\
\midrule
Greedy (ours, $\varepsilon{=}0$)           & \textbf{0.4308}$\pm$.0064 & \textbf{59.3}$\pm$7.0 \\
$\varepsilon$-greedy ($\varepsilon{=}0.1$) & 0.4156$\pm$.0069          & 70.7$\pm$10.8 \\
$\varepsilon$-greedy ($\varepsilon{=}0.2$) & 0.4162$\pm$.0036          & 77.7$\pm$1.2  \\
\midrule
25\% meta-datasets & 0.3358$\pm$.0079 & 109.0$\pm$0.0 \\
50\% meta-datasets & 0.3743$\pm$.0053 &  96.0$\pm$3.5 \\
75\% meta-datasets & 0.4122$\pm$.0013 &  92.0$\pm$14.0 \\
100\% (ours)       & \textbf{0.4308}$\pm$.0064 & \textbf{59.3}$\pm$7.0 \\
\bottomrule
\end{tabular}
\end{table}

Adding exploration ($\varepsilon > 0$) degrades both AP and rank, as random steps create a train/test distribution mismatch---the deployed policy is purely greedy, so training should match this structure. Meta-dataset coverage improves monotonically, confirming that diversity across meta-tasks is critical for transferable gain prediction.

\subsection{State Representation Features}
\label{app:subsec:features}

\paragraph{State Representation.}
The state representation $\phi(f_i, f^*_{i-1}, P)$ is a fixed 61-dimensional feature vector that encodes the interaction between a candidate model $f_i$, the most recently selected detector $f^*_{i-1}$, and the current ensemble $P$. 
It consists of three 20-dimensional feature blocks and one scalar feature, namely:
(i) $\phi_{f^*_{i-1}, f_{i}}$, capturing pairwise interactions between the last selected model and the candidate;
(ii) ${\phi}_{f^{*}_{i-1}, P}$, summarizing interactions between the last selected model and the existing ensemble;
(iii) $\phi_{f_{i}, P}$, summarizing interactions between the ensemble and the candidate; and
(iv) $|P|$, the current ensemble cardinality.

Each 20-dimensional block is composed of 15 base pairwise features and 5 context features, computed from normalized outlier score vectors $\mathbf{o}_{f} \in [0,1]^N$, where scores are Min--Max normalized independently for each detector and dataset. 
Table~\ref{app:tab:features} provides the complete specification of all feature dimensions.

\begin{table*}[t]
\centering
\caption{Feature specification for pairwise model comparison. All features are computed from normalized outlier score vectors $\mathbf{o}_p, \mathbf{o}_q \in [0,1]^N$.}
\label{app:tab:features}
\fontsize{8}{8}\selectfont
\begin{tabular}{@{}lp{2.5cm}p{10.5cm}@{}}
\toprule
\textbf{Feature} & \textbf{Type} & \textbf{Mathematical Definition} \\
\midrule
\multicolumn{3}{l}{\textit{Base Pairwise Features (15 dimensions)}} \\
\midrule
\texttt{pearson} & Correlation & Pearson correlation: $\rho(\mathbf{o}_p, \mathbf{o}_q) = \frac{\text{cov}(\mathbf{o}_p, \mathbf{o}_q)}{\sigma_p \sigma_q}$ \\
\texttt{tail\_pearson} & Correlation & Pearson correlation on $\mathcal{U} = \text{top-}k_p \cup \text{top-}k_q$ where $k = \lceil 0.1N \rceil$ \\
\texttt{jaccard} & Set similarity & Jaccard index: $J = |\text{top-}k_p \cap \text{top-}k_q| / |\text{top-}k_p \cup \text{top-}k_q|$ \\
\texttt{rel\_kurtosis} & Moment ratio & Relative kurtosis: $\text{kurt}(\mathbf{o}_q) / \max(\text{kurt}(\mathbf{o}_p), 10^{-3})$ \\
\texttt{tail\_pos\_disagreement} & Distributional & Mean positive difference: $\mathbb{E}_{i \in \mathcal{U}}[\max(0, \mathbf{o}_{q,i} - \mathbf{o}_{p,i})]$ \\
\texttt{centrality} & Distributional & Correlation with pool mean: $\rho(\mathbf{o}_q, \bar{\mathbf{o}})$ where $\bar{\mathbf{o}} = \frac{1}{K}\sum_{j=1}^K \mathbf{o}_j$ \\
\texttt{cand\_std} & Distributional & Standard deviation: $\sigma(\mathbf{o}_q)$ \\
\texttt{pseudo\_ap} & Pseudo-label & Average Precision treating $\text{top-}k_p$ as positive class, $\mathbf{o}_q$ as predictions \\
\texttt{pseudo\_roc} & Pseudo-label & ROC AUC treating $\text{top-}k_p$ as positive class, $\mathbf{o}_q$ as predictions \\
\texttt{tail\_entropy} & Distributional & Shannon entropy: $H(\mathbf{o}_q) = -\sum_{b=1}^{10} p_b \log p_b$ where $p_b$ is 10-bin histogram \\
\texttt{score\_dist\_l2} & Distance & $L_2$ distance of sorted vectors: $\|\text{sort}(\mathbf{o}_p) - \text{sort}(\mathbf{o}_q)\|_2$ \\
\texttt{cosine\_dist} & Distance & Cosine distance: $1 - \frac{\mathbf{o}_p \cdot \mathbf{o}_q}{\|\mathbf{o}_p\|_2 \|\mathbf{o}_q\|_2}$ \\
\texttt{tail\_divergence} & Distance & Mean absolute difference: $\mathbb{E}_{i \in \mathcal{U}}[|\mathbf{o}_{q,i} - \mathbf{o}_{p,i}|]$ \\
\texttt{same\_family} & Categorical & Binary indicator: $\mathbb{I}[\mathcal{F}(p) = \mathcal{F}(q)]$ where $\mathcal{F}$ maps models to families \\
\texttt{same\_as\_ensemble\_count} & Set overlap & Count of partners $s \in S$ with $J(q, s) > 0.5$; equals 0 when $S = \emptyset$ \\
\midrule
\multicolumn{3}{l}{\textit{Primary-Context Features (5 dimensions)}} \\
\midrule
\texttt{prim\_std} & Distributional & Standard deviation of primary: $\sigma(\mathbf{o}_p)$ \\
\texttt{prim\_entropy} & Distributional & Shannon entropy of primary score distribution: $H(\mathbf{o}_p)$ \\
\texttt{prim\_centrality} & Distributional & Centrality of primary: $\rho(\mathbf{o}_p, \bar{\mathbf{o}})$ \\
\texttt{prim\_kurtosis} & Moment & Excess kurtosis of primary: $\text{kurt}(\mathbf{o}_p)$ \\
\texttt{prim\_skewness} & Moment & Skewness of primary: $\text{skew}(\mathbf{o}_p)$ \\
\midrule
\multicolumn{3}{l}{\textit{Set-Level Scalar Features (1 dimension)}} \\
\midrule
$|P|$ & Cardinality & Number of partners: $|S| \in \{0, 1, 2, \ldots\}$ \\
\bottomrule
\end{tabular}
\end{table*}

\paragraph{Feature Aggregation for Ensemble Context.}
For ensembles with $|P| > 1$, set-level context features are computed via mean pooling to ensure permutation invariance:
\begin{align*}
\phi_{f^*_{i-1},P} &= \frac{1}{|P|-1} \sum_{f \in P \setminus \{f^*_{i-1}\}} \phi_{f^*_{i-1}, f} \\
\phi_{f_{i},P} &= \frac{1}{|P|} \sum_{f \in P} \phi_{f,f_i}
\end{align*}
Here, the first equation measures the variance between the last selected model and the current ensemble. The second equation measures the variance between the candidate model and the current ensemble.
When $|P| = 1$ (i.e., only the primary detector has been selected), $\phi_{f^*_{i-1},P}$ is zero-padded, while $\phi_{f_{i},P}$ is computed directly from the primary detector. 
This design maintains a fixed 61-dimensional representation across all selection stages and allows the meta-model to distinguish early-stage expansion from later iterative selection.

\subsection{Effects of Meta-Model Architectures}
\label{app:subsec:meta_model_analysis}

The effectiveness of \framework depends on the underlying models used to predict marginal gains ($f_{\text{reg}}$) and filter candidates ($f_{\text{cls}}$). In this section, we justify our default choice of \texttt{ExtraTrees} by comparing it with five alternative learning families: \texttt{Random Forest}~\cite{DBLP:journals/ml/Breiman01}, \texttt{XGBoost}~\cite{DBLP:conf/kdd/ChenG16}, \texttt{LightGBM}~\cite{DBLP:conf/nips/KeMFWCMYL17}, \texttt{Multi-Layer Perceptrons}~\cite{DBLP:book/GoodfellowBC2016}, and \texttt{Linear Models}~\cite{DBLP:book/JamesWHT13}. All methods are evaluated using the same meta-features and training protocol.

To ensure a fair comparison, we employ strong, standardized configurations across baselines. Tree ensembles (\texttt{Random Forest}, \texttt{ExtraTrees}, \texttt{XGBoost}, \texttt{LightGBM}) use 500 estimators for classification tasks and 800 for regression tasks. Gradient boosting models (\texttt{XGBoost}, \texttt{LightGBM}) adopt a subsample ratio of 0.8 to improve generalization. The \texttt{Multi-Layer Perceptrons} model consists of two hidden layers with sizes $(100, 50)$, trained for 500 epochs. \texttt{Linear Models} use Logistic Regression and Ridge Regression with default regularization.

\begin{table*}[h]
\centering
\caption{Comparison of meta-model architectures for \framework selection quality.}
\small
\label{app:tab:meta_model_comparison}
\begin{tabular}{@{}llllll@{}}
\toprule
\textbf{Method} & \textbf{AP (Mean $\pm$ Std)} $\uparrow$ & \textbf{Rank} $\downarrow$ & \textbf{ROC-AUC} $\uparrow$ & \textbf{Train (s)} $\downarrow$ & \textbf{Infer (s)} $\downarrow$ \\
\midrule
\texttt{ExtraTrees} (\textbf{Ours}) & \textbf{0.4308} $\pm$ 0.0064 & \textbf{59.3} $\pm$ 6.9610 & \textbf{0.7867} $\pm$ 0.0045 & 4.1518 & 1.6839 \\
\texttt{Linear Models} & 0.4056 $\pm$ 0.0015 & 105.2 $\pm$ 2.5000 & 0.7737 $\pm$ 0.0010 & 6.9150 & \textbf{1.0327} \\
\texttt{XGBoost}            & 0.4110 $\pm$ 0.0055 & 108.0 $\pm$ 8.2000 & 0.7664 $\pm$ 0.0040 & \textbf{1.7408} & 1.4910 \\
\texttt{Random Forest}      & 0.4043 $\pm$ 0.0062 & 77.0 $\pm$ 7.1000 & 0.7617 $\pm$ 0.0050 & 15.5561 & 1.5200 \\
\texttt{LightGBM}           & 0.3899 $\pm$ 0.0058 & 99.0 $\pm$ 8.5000 & 0.7600 $\pm$ 0.0042 & 1.9412 & 1.3405 \\
\texttt{Multi-Layer Perceptrons}                & 0.3711 $\pm$ 0.0115 & 110.0 $\pm$ 12.5000 & 0.7607 $\pm$ 0.0090 & 35.6667 & 10.7667 \\
\bottomrule
\end{tabular}%
\end{table*}

Table~\ref{app:tab:meta_model_comparison} summarizes the results. \textit{Train (s)} denotes the average wall-clock time to train the meta-model on the meta-training set (leave-one-out protocol), while \textit{Infer (s)} measures the time required to select models for a new dataset. \texttt{ExtraTrees} consistently achieves the best predictive performance, with the highest AP ($0.4308$) and best average rank ($59.3$), while maintaining competitive training ($4.15$s) and inference ($1.68$s). \texttt{XGBoost} provides faster training but trails in AP ($0.4110$) and rank ($108.0$), suggesting overfitting under the relatively small meta-training regime. \texttt{Linear Models} offer the fastest inference ($1.03$s) but cannot capture the non-linear relationships required for accurate selection, resulting in substantially worse ranking performance. \texttt{Random Forest} is competitive in rank but is slower to train and less accurate than \texttt{ExtraTrees}; this supports the benefit of fully randomized splits as additional regularization. \texttt{Multi-Layer Perceptrons} perform poorly both in accuracy (lowest AP $0.3711$) and computational efficiency, indicating that neural architectures are less suitable for this structured meta-learning task. Overall, the comparison shows that \texttt{ExtraTrees} provides the best accuracy--efficiency tradeoff for our two-part marginal-gain predictor.

\subsection{Framework Algorithm}
\label{app:subsec:algorithm}

\paragraph{Offline Meta-training Procedure.}
Algorithm~\ref{alg:offline_meta_training_algx} summarizes the offline meta-training phase of \framework. 
Given labeled meta-datasets $\mathcal{M}=\{(\mathbf{M}_\ell,\mathbf{y}_\ell)\}$ and a detector pool $\Omega$, we simulate oracle greedy ensemble construction by iteratively expanding an ensemble $P \subseteq \Omega$. 
Starting from an oracle-selected primary detector $f^*_1$, at each step we evaluate every candidate $f_i \in \Omega \setminus P$ using its state representation $\phi(f_i, f^*_{i-1}, P)$ and compute the true marginal gain $G(f_i \mid P)$ in Average Precision. 
These state–gain pairs are used to train a two-part meta-model that estimates the probability and magnitude of marginal improvement, yielding a predictor $\hat{G}(f_i \mid P)$ that is later used to guide label-free online ensemble selection.

\begin{algorithm}[t]
\caption{\framework Offline Meta-training: Oracle Rollouts for Marginal-Gain Supervision}
\label{alg:offline_meta_training_algx}
\footnotesize
\begin{algorithmic}[1]
\REQUIRE Labeled meta-datasets $\mathcal{M}=\{(\mathbf{M}_\ell,\mathbf{y}_\ell)\}_{\ell=1}^{L}$; detector pool $\Omega$; budget $\eta$; rollout length $T \le \eta$; state function $\phi(f_i, f^*_{i-1}, P)$
\ENSURE Trained meta-models $(f_{\text{cls}}, f_{\text{reg}})$

\STATE Initialize $\mathcal{D}_{\text{cls}} \leftarrow \emptyset$, $\mathcal{D}_{\text{reg}} \leftarrow \emptyset$
\FORALL{$(\mathbf{M}, \mathbf{y}) \in \mathcal{M}$}
    \STATE \textbf{(Oracle primary)} 
    \STATE $f^*_1 \leftarrow \arg\max_{f \in \Omega}\; \mathrm{AP}(f(\mathbf{M}), \mathbf{y})$
    \STATE $P \leftarrow \{f^*_1\}$ 
    \STATE $f_{\text{last}} \leftarrow f^*_1$
    \FOR{$t = 2$ to $T$}
        \FORALL{$f_i \in \Omega \setminus P$}
            \STATE $\boldsymbol{\phi}_i \leftarrow \phi(f_i, f_{\text{last}}, P)$
            \STATE $G(f_i \mid P) \leftarrow \mathrm{AP}(\mathbf{o}_{P \cup \{f_i\}}, \mathbf{y}) - \mathrm{AP}(\mathbf{o}_{P}, \mathbf{y})$
            \STATE $y_{\text{cls}} \leftarrow \mathbb{I}(G(f_i \mid P) > 0)$
            \STATE $\mathcal{D}_{\text{cls}} \leftarrow \mathcal{D}_{\text{cls}} \cup \{(\boldsymbol{\phi}_i, y_{\text{cls}})\}$
            \IF{$y_{\text{cls}} = 1$}
                \STATE $y_{\text{reg}} \leftarrow G(f_i \mid P)$
                \STATE $\mathcal{D}_{\text{reg}} \leftarrow \mathcal{D}_{\text{reg}} \cup \{(\boldsymbol{\phi}_i, y_{\text{reg}})\}$
            \ENDIF
        \ENDFOR
        \STATE $f^* \leftarrow \arg\max_{f_i \in \Omega \setminus P}\; G(f_i \mid P)$
        \IF{$G(f^* \mid P) \le 0$}
            \STATE \textbf{break}
        \ENDIF
        \STATE $P \leftarrow P \cup \{f^*\}$; $f_{\text{last}} \leftarrow f^*$
        \IF{$|P| \ge \eta$}
            \STATE \textbf{break}
        \ENDIF
    \ENDFOR
\ENDFOR
\STATE Train classifier $f_{\text{cls}}$ on $\mathcal{D}_{\text{cls}}$ (balanced weights)
\STATE Train regressor $f_{\text{reg}}$ on $\mathcal{D}_{\text{reg}}$ (positives only)
\STATE \textbf{return} $(f_{\text{cls}}, f_{\text{reg}})$
\end{algorithmic}
\end{algorithm}

\paragraph{Online Model Selection Procedure.}
Algorithm~\ref{alg:online_model_selection_algx} describes the online ensemble selection procedure of \framework for a new unlabeled dataset. 
Starting from a primary detector $f^*_1$ selected by an unsupervised criterion, we iteratively construct an ensemble $P$ by adding candidates $f_i \in \Omega \setminus P$ that maximize the proxy marginal utility $\Delta U(f_i \mid P)$. 
The proxy utility combines the predicted marginal gain $\hat{G}(f_i \mid P)$ from the meta-model with redundancy discounting and family-risk regularization, thereby enforcing diminishing returns as the ensemble grows. 
The selection process terminates automatically when $\max_{f_i} \Delta U(f_i \mid P) \le 0$ or when the budget $|P|=\eta$ is reached, yielding compact, dataset-adaptive ensembles without access to labels.

\begin{algorithm}[t]
\caption{\framework: Online Selection via Submodular-Inspired Proxy Maximization}
\label{alg:online_model_selection_algx}
\footnotesize
\begin{algorithmic}[1]
\REQUIRE dataset $\mathbf{X}$, candidate pool $\Omega$, meta-model $M(\cdot)$ producing $\hat{G}$, state representation $\phi(\cdot)$, family mapping $\mathcal{F}$, thresholds $\tau_1,\tau_2$, budget $\eta \ge 1$, diversity $\beta$, risk $\lambda_{\text{fam}}$, $k_{\text{top}}$
\ENSURE Ensemble $P$

\STATE \textit{Primary model selection using \texttt{ELECT} as $\mathcal{S}(\cdot)$}
\STATE $f^*_1 \leftarrow \mathcal{S}(\mathbf{X}, \Omega)$
\STATE $P \leftarrow \{f^*_1\}$
\STATE $\Omega_{\text{pool}} \leftarrow \Omega \setminus \{f^*_1\}$

\STATE \textit{Stage 1: Select first expansion model}
\FORALL{$f_i \in \Omega_{\text{pool}}$}
    \STATE $\hat{G}(f_i \mid P) \leftarrow M.\text{predict}\!\left(\phi(f_i, f^*_1, P)\right)$
\ENDFOR
\STATE $f^*_2 \leftarrow \arg\max_{f_i \in \Omega_{\text{pool}}}\; \hat{G}(f_i \mid P)$
\IF{$\hat{G}(f^*_2 \mid P) < \tau_1$}
    \STATE \textbf{return} $P$
\ENDIF
\STATE $P \leftarrow P \cup \{f^*_2\}$

\STATE \textit{Stage 2: Iterative expansion}
\WHILE{$|P| < \eta$}
    \STATE $u_{\text{best}} \leftarrow 0$
    \FORALL{$f_i \in \Omega_{\text{pool}} \setminus P$}
        \STATE $\hat{G}(f_i \mid P) \leftarrow M.\text{predict}\!\left(\phi(f_i, f^*_{|P|-1}, P)\right)$
        \IF{$\hat{G}(f_i \mid P) < \tau_2$}
            \STATE \textbf{continue}
        \ENDIF
        \STATE Compute $\Delta U(f_i \mid P)$
        \IF{$\Delta U(f_i \mid P) > u_{\text{best}}$}
            \STATE $f^* \leftarrow f_i$
            \STATE $u_{\text{best}} \leftarrow \Delta U(f_i \mid P)$
        \ENDIF
    \ENDFOR
    \IF{$u_{\text{best}} \le 0$}
        \STATE \textbf{break}
    \ENDIF
    \STATE $P \leftarrow P \cup \{f^*\}$
\ENDWHILE
\STATE \textbf{return} $P$
\end{algorithmic}
\end{algorithm}

\subsection{Complexity Analysis}
\label{app:subsec:complexity_analysis}

We conduct a computational complexity analysis to clarify how the proposed framework scales with the size of the candidate model pool and dataset size.

\framework operates in $O(|\Omega|^2 \cdot N + \eta |\Omega| \cdot C_{\text{inf}})$ time, where $|\Omega|$ is the size of the candidate model pool, $N$ is the number of samples in the target dataset, $\eta$ is the ensemble budget, and $C_{\text{inf}}$ is the meta-model inference cost. 
The $O(|\Omega|^2 \cdot N)$ term corresponds to one-time feature pre-computation of similarity statistics for the target dataset: correlation-based features require $O(|\Omega|^2 \cdot N)$ operations, while Jaccard similarity over top-$k_{\text{top}}$ instances requires $O(|\Omega| \cdot N)$ time for extracting top-$k_{\text{top}}$ indices and $O(|\Omega|^2 \cdot k_{\text{top}})$ time for set intersections, becoming independent of $N$ after ranking. 
The interactive selection phase has complexity $O(\eta |\Omega| \cdot C_{\text{inf}})$ and is independent of $N$. 
For tree-based meta-models, $C_{\text{inf}} = O(n_{\text{trees}} \cdot d_\phi)$, where $n_{\text{trees}}$ is the number of trees and $d_\phi$ is the feature dimensionality.

\subsection{Datasets}
\label{app:subsec:datasets}

We conduct experiments on the 39-dataset benchmark introduced in \texttt{ELECT}~\cite{DBLP:conf/icdm/0016ZA22}, which comprises independent, real-world tabular anomaly detection tasks sourced from the Outlier Detection Data Sets (ODDS)\footnote{https://giftpathao.com}
 and the UCI Machine Learning Repository~\cite{DBLP:journals/datamine/CamposZSCMSAH16}. Detailed characteristics of all datasets used in our experiments are summarized in Table~\ref{app:tab:datasets}.
 
\begin{table*}[t]
    \centering
    \caption{Summary of benchmark datasets.}
    \label{app:tab:datasets}
    \fontsize{8}{8}\selectfont
    \begin{tabular}{lrrrl}
        \toprule
        \textbf{Dataset} & \textbf{\#Samples} & \textbf{Dimensionality} & \textbf{Contamination} (\%) & \textbf{Type} \\
        \midrule
        ALOI              & 49,534 & 27  &  3.04 & Numeric \\
        Annthyroid        & 7,129  & 21  &  7.49 & Mixed \\
        Arrhythmia        & 450   & 259 & 45.78 & Mixed \\
        Cardiotocography  & 2,114  & 21  & 22.04 & Numeric \\
        Glass             & 214   & 7   &  4.21 & Numeric \\
        HeartDisease      & 270   & 13  & 44.44 & Mixed \\
        InternetAds       & 1,966  & 1,555 & 18.72 & Mixed \\
        PageBlocks        & 5,393  & 10  &  9.46 & Numeric \\
        PenDigits         & 9,868  & 16  &  0.20 & Numeric \\
        Pima              & 768   & 8   & 34.90 & Numeric \\
        Shuttle           & 1,013  & 9   &  1.28 & Numeric \\
        SpamBase          & 4,207  & 57  & 39.91 & Numeric \\
        Stamps            & 340   & 9   &  9.12 & Numeric \\
        WBC               & 223   & 9   &  4.48 & Numeric \\
        WDBC              & 367   & 30  &  2.72 & Categorical \\
        WPBC              & 198   & 33  & 23.74 & Numeric \\
        Waveform          & 3,443  & 21  &  2.90 & Numeric \\
        Wilt              & 4,819  & 5   &  5.33 & Numeric \\
        annthyroid        & 7,200  & 6   &  7.42 & Mixed \\
        arrhythmia        & 452   & 274 & 14.60 & Mixed \\
        breastw           & 683   & 9   & 34.99 & Numeric \\
        glass             & 214   & 9   &  4.21 & Numeric \\
        ionosphere        & 351   & 33  & 35.90 & Numeric \\
        letter            & 1,600  & 32  &  6.25 & Numeric \\
        lympho            & 148   & 18  &  4.05 & Categorical \\
        mammography       & 11,183 & 6   &  2.32 & Numeric \\
        mnist             & 7,603  & 100 &  9.21 & Numeric \\
        musk              & 3,062  & 166 &  3.17 & Numeric \\
        optdigits         & 5,216  & 64  &  2.88 & Numeric \\
        pendigits         & 6,870  & 16  &  2.27 & Numeric \\
        pima              & 768   & 8   & 34.90 & Numeric \\
        satellite         & 6,435  & 36  & 31.64 & Numeric \\
        satimage-2        & 5,803  & 36  &  1.22 & Numeric \\
        speech            & 3,686  & 400 &  1.65 & Numeric \\
        thyroid           & 3,772  & 6   &  2.47 & Mixed \\
        vertebral         & 240   & 6   & 12.50 & Numeric \\
        vowels            & 1,456  & 12  &  3.43 & Numeric \\
        wbc               & 378   & 30  &  5.56 & Categorical \\
        wine              & 129   & 13  &  7.75 & Numeric \\
        \bottomrule
    \end{tabular}
\end{table*}

\subsection{Candidate Model Pool}
\label{app:subsec:model_pool}

We construct a diverse candidate pool of $M = 297$ unsupervised outlier detection models by systematically varying hyperparameters within 8 widely-used algorithmic families. For each dataset, all 297 models are pre-fitted and their anomaly scores cached for efficient meta-learning experiments. Table~\ref{app:tab:model_pool} summarizes the configuration. 

\begin{table*}[t]
\centering
\caption{Candidate model pool specification ($M = 297$ total models).}
\label{app:tab:model_pool}
\centering
\small
\begin{tabular}{llp{7cm}}
\toprule
\textbf{Family} & \textbf{Count} & \textbf{Hyperparameter Grid} \\
\midrule
\texttt{kNN}~\cite{DBLP:conf/sdm/Chehreghani16} & 36 & \makecell{method $\in \{\text{largest}, \text{mean}, \text{median}\}$;\\ $k \in \{1, 5, 10, 15, 20, 25, 50, 60, 70, 80, 90, 100\}$} \\
\midrule
\texttt{LOF}~\cite{DBLP:conf/sigmod/BreunigKNS00} & 36 & \makecell{metric $\in \{\text{euclidean}, \text{manhattan}, \text{minkowski}\}$; \\$k \in \{1, 5, 10, 15, 20, 25, 50, 60, 70, 80, 90, 100\}$} \\
\midrule
\texttt{IForest}~\cite{DBLP:conf/icdm/LiuTZ08} & 81 & \makecell{$n_{\text{estimators}} \in \{10, 20, 30, 40, 50, 75, 100, 150, 200\}$; \\max\_samples $\in \{0.1, 0.2, \ldots, 0.9\}$} \\
\midrule
\texttt{HBOS}~\cite{goldstein2012histogram} & 40 & \makecell{$n_{\text{bins}} \in \{5, 10, 20, 30, 40, 50, 75, 100\}$; \\tolerance $\in \{0.1, 0.2, 0.3, 0.4, 0.5\}$} \\
\midrule
\texttt{OCSVM}~\cite{DBLP:conf/nips/ScholkopfWSSP99} & 36 & \makecell{kernel $\in \{\text{linear}, \text{poly}, \text{rbf}, \text{sigmoid}\}$; \\$\nu \in \{0.1, 0.2, \ldots, 0.9\}$} \\
\midrule
\texttt{LODA}~\cite{DBLP:journals/ml/Pevny16} & 54 & \makecell{$n_{\text{bins}} \in \{5, 10, 15, 20, 25, 30\}$; \\$n_{\text{cuts}} \in \{10, 20, 30, 40, 50, 75, 100, 150, 200\}$} \\
\midrule
\texttt{ABOD}~\cite{DBLP:conf/kdd/KriegelSZ08} & 7 & $n_{\text{neighbors}} \in \{3, 5, 10, 15, 20, 25, 50\}$ \\
\midrule
\texttt{COF}~\cite{DBLP:conf/pakdd/TangCFC02} & 7 & $n_{\text{neighbors}} \in \{3, 5, 10, 15, 20, 25, 50\}$ \\
\midrule
\textbf{Total} & \textbf{297} & \\
\bottomrule
\end{tabular}
\end{table*}

\subsection{Baselines}
\label{app:subsec:baselines}
We compare the proposed \framework framework against 19 unsupervised baselines categorized into four groups. For ensemble baselines dependent on a size parameter $k$, we report the best $k$ performance---the fixed ensemble size that yields the highest average AP across the benchmark---to ensure a strong competitive baseline. 

\textbf{Single Baselines:} 
(1) \texttt{Singleton}: Selects a single model uniformly at random from the candidate pool;
(2) \texttt{LOF}~\cite{DBLP:conf/sigmod/BreunigKNS00}; 
(3) \texttt{Global Best}: A static meta-learner selecting the single model with the highest average performance across all training datasets; 

\textbf{Na\"ive \& Random Ensemble Baselines:}
(4) \texttt{IForest}~\cite{DBLP:conf/icdm/LiuTZ08}: A tree-based ensemble that isolates anomalies using random partitions; 
(5) \texttt{RandNet}~\cite{DBLP:conf/sdm/ChenSAT17}: An ensemble of autoencoders with randomized connectivity to mitigate overfitting and enhance diversity;
(6) \texttt{Random Ensemble}: An ensemble of $k$ randomly selected models; 
(7) \texttt{Mega Ensemble}: A na\"ive ensemble that averages outlier scores from all 297 models in the pool. 

\textbf{Deep Learning Baselines:} (8) \texttt{RDA}~\cite{DBLP:conf/kdd/ZhouP17}: A robust deep autoencoder that decomposes input into low-dimensional manifold and sparse noise components;
(9) \texttt{DAGMM}~\cite{DBLP:conf/iclr/ZongSMCLCC18}: An end-to-end framework jointly optimizing a compression network and a Gaussian Mixture Model for density estimation;
(10) \texttt{DeepSVDD}~\cite{DBLP:conf/icml/RuffGDSVBMK18}: Maps data into a minimum volume hypersphere to extract common factors of variation;
(11) \texttt{ROBOD}~\cite{DBLP:conf/nips/DingZA22}: A deep hyper-ensemble that aggregates multiple MLP detectors trained under varied hyperparameters and unsupervised contamination assumptions;
(12) \texttt{LUNAR}~\cite{DBLP:conf/aaai/GoodgeHNN22}: A graph-based contrastive representation learner for tabular anomaly detection;
(13) \texttt{DTE-C}~\cite{DBLP:conf/iclr/LivernocheJHR24}: A diffusion-based generative model for tabular anomaly detection with a transductive scoring protocol aligned to our benchmark;
(14) \texttt{TCCM}~\cite{DBLP:journals/corr/abs-2510-18328}: A flow-matching generative model for tabular anomaly detection, evaluated with the same protocol as \texttt{DTE-C}.

\textbf{Meta-Learning and Ensemble Baselines:}
(15) \texttt{MetaOD}~\cite{DBLP:journals/corr/abs-2009-10606}: A model selection as a cold-start recommendation problem using matrix factorization on historical meta-features;
(16) \texttt{LSCP}~\cite{DBLP:conf/sdm/ZhaoNHL19}: An ensemble framework that selects competent base detectors for local regions of test instances;
(17) \texttt{ELECT}~\cite{DBLP:conf/icdm/0016ZA22}: The current state-of-the-art model selector, which identifies the single best model based on performance-driven task similarity.
(18) \texttt{ELECT+k}: An ensemble baseline that aggregates the top-$k$ models ranked by \texttt{ELECT}. This tests whether simple rank-based selection is sufficient compared to our context-aware approach.
(19) \texttt{ELECT+r}: A hybrid baseline that initializes with the high-quality primary model selected by \texttt{ELECT} but expands the ensemble with random partners. 

\textbf{Theoretical Upper Bound:}
(20) \texttt{Greedy Oracle}: A fully supervised upper bound that utilizes ground-truth labels to iteratively select the candidate maximizing marginal AP at each step. This baseline quantifies the maximum potential performance achievable by a greedy sequential selection strategy.

\subsection{Implementation Details}
\label{app:subsec:implementation}
We implement the proposed framework using PyTorch 1.13~\cite{DBLP:conf/nips/PaszkeGMLBCKLGA19} for neural components and Scikit-learn 1.2~\cite{DBLP:journals/jmlr/PedregosaVGMTGBPWDVPCBPD11} and PyOD 1.0~\cite{DBLP:journals/jmlr/ZhaoNL19} for base outlier detectors in Python 3.9. All experiments were executed on a high-performance computing cluster equipped with dual NVIDIA RTX 3090 GPUs (24GB VRAM), AMD EPYC 7742 64-Core Processors, and 512GB RAM. The source code is available at \underline{\url{https://anonymous.4open.science/r/MetaEns}}. For preprocessing, numeric features are scaled using RobustScaler (median removal and scaling according to the interquartile range) to preserve the integrity of global outliers. Categorical features are transformed via target encoding with leave-one-out regularization ($\alpha=10$) to prevent data leakage. We add binary missingness indicators and impute missing values with median/mode. Since raw anomaly scores from different algorithms operate on vastly different scales, we strictly apply Min-Max Normalization per model per dataset to all model outputs before aggregation. To ensure reproducibility, we fix random seeds to $42$ for all stochastic components.

\subsection{Evaluation Metrics}
\label{app:metrics}

Let $\{(\mathbf{x}_i, y_i)\}_{i=1}^{N}$ denote an evaluation dataset, where $y_i \in \{0,1\}$ indicates whether instance $\mathbf{x}_i$ is anomalous ($y_i=1$) or normal ($y_i=0$). 
Let $\mathbf{o} = (o_1,\dots,o_N)$ denote the outlier score vector produced by a method, where larger values indicate higher anomaly likelihood. 
Let $\pi = \sum_{i=1}^{N} y_{i}$ denote the number of anomalies in the dataset. 
All ranking-based metrics are computed by sorting instances in descending order of $o_i$.

\paragraph{Average Precision (AP).}
Average Precision summarizes the area under the precision--recall curve:
\begin{equation}
\mathrm{AP} = \frac{1}{\pi} \sum_{k=1}^{N} \mathbb{I}\!\left(y_{(k)} = 1\right)\, P(k),
\end{equation}
where $y_{(k)}$ is the label of the instance ranked at position $k$, and $P(k)$ is the precision at cutoff $k$:
\begin{equation}
P(k) = \frac{1}{k} \sum_{j=1}^{k} \mathbb{I}\!\left(y_{(j)} = 1\right)
\end{equation}

\paragraph{Average Rank (AR).}
Let $r_{f_{i},\mathbf{X}_{j}}$ denote the rank of method $f_{i}$ on dataset $\mathbf{X}_{j}$ according to Average Precision, where rank $1$ indicates the best-performing method. 
The Average Rank of method $f_{i}$ is defined as follows.
\begin{equation}
\mathrm{AR}_{f_{i}} = \frac{1}{|\mathcal{D}|} \sum_{\mathbf{X}_{j} \in \mathcal{D}} r_{f_{i},\mathbf{X}_{j}},
\end{equation}
where $\mathcal{D}$ denotes the set of benchmark datasets. Lower values indicate better overall performance.

\paragraph{ROC-AUC.}
The ROC-AUC measures the probability that an anomalous instance receives a higher outlier score than a normal instance:
\begin{equation}
\mathrm{AUC} =
\mathbb{P}(o^+ > o^-)
= \frac{\sum_{i: y_i = 1} \sum_{j: y_j = 0} \mathbb{I}(o_i > o_j)}{\pi (N - \pi)},
\end{equation}
where $o^+$ and $o^-$ denote outlier scores of anomalous and normal instances, respectively.

\paragraph{Precision@$\pi$.}
Precision@$\pi$ evaluates the accuracy of the top-ranked predictions at the exact number of ground-truth anomalies and is defined as
\begin{equation}
\mathrm{Precision@}\pi = \frac{1}{\pi} \sum_{k=1}^{\pi} \mathbb{I}\!\left(y_{(k)} = 1\right),
\label{eq:precision_pi}
\end{equation}
where $\pi=\sum_{i=1}^{N}y_i$ is the number of anomalous instances in the dataset and $y_{(k)}$ is the label of the instance ranked at position $k$. This metric reflects practical inspection scenarios where only the top-ranked instances are examined.

\paragraph{Max F1-score.}
For a decision threshold $\tau$ applied to outlier scores, predicted labels are defined as follows.
\begin{equation}
\hat{y}_i(\tau) = \mathbb{I}(o_i \ge \tau)
\end{equation}
Precision and recall at threshold $\tau$ are given as follows.
\begin{equation}
\begin{aligned}
\mathrm{Prec}(\tau) = \frac{\sum_i \mathbb{I}(\hat{y}_i(\tau)=1 \wedge y_i=1)} {\sum_i \mathbb{I}(\hat{y}_i(\tau)=1)}\\
\mathrm{Rec}(\tau) = \frac{\sum_i \mathbb{I}(\hat{y}_i(\tau)=1 \wedge y_i=1)} {\sum_i \mathbb{I}(y_i=1)}
\end{aligned}
\end{equation}
The F1-score at threshold $\tau$ is defined as follows.
\begin{equation}
\mathrm{F1}(\tau) =
\frac{2 \cdot \mathrm{Prec}(\tau)\cdot \mathrm{Rec}(\tau)}
{\mathrm{Prec}(\tau)+\mathrm{Rec}(\tau)}
\end{equation}
The Max F1-score is defined as follows.
\begin{equation}
\mathrm{MaxF1} = \max_{\tau} \mathrm{F1}(\tau)
\end{equation}

\subsection{Detailed Experimental Results}
\label{app:sec:detailed_experiments}
Table~\ref{tab:detailed_performance} reports the detailed performance of \framework and all baselines on each of the 39 benchmark datasets. 
For each dataset, we report Average Precision (AP) scores computed under the same evaluation protocol described in Section~\ref{sec:experiments}. 
This table complements the aggregated results in the main paper by providing a per-dataset view of method behavior, enabling fine-grained comparison and reproducibility analysis.

\begin{table*}[t]
\centering
\caption{Detailed Performance Comparison Across 39 Benchmark Datasets. For each method, we report Average Precision (AP) and rank in parentheses (lower rank is better, 1--19, among compared methods). Best results in \textbf{bold}. Per-dataset values are from a single representative seed (seed=42); the 3-seed averaged AP for \framework is \textbf{0.4308} (Table~\ref{tab:baselines}). Abbreviations: \texttt{RS} = \texttt{Random Selection}, RE = \texttt{Random Ensemble} (k=3), \texttt{ELECT} = \texttt{ELECT} (Top-1), \texttt{ELECT-10} = \texttt{ELECT} (Top-10).}
\label{tab:detailed_performance}
\small
\setlength{\tabcolsep}{3pt}
\resizebox{\textwidth}{!}{
\begin{tabular}{l|llllllllllllllllll|l}
\toprule
\textbf{Dataset} & \textbf{RS} & \textbf{IForest} & \textbf{LOF} & \textbf{GB} & \textbf{ME} & \textbf{RE} & \textbf{RDA} & \textbf{DAGMM} & \textbf{DeepSVDD} & \textbf{RandNet} & \textbf{ROBOD} & \textbf{LUNAR} & \textbf{DTE-C} & \textbf{TCCM} & \textbf{LSCP} & \textbf{MetaOD} & \textbf{ELECT-1} & \textbf{ELECT-10} & \textbf{MetaEns} \\
\midrule
ALOI & 0.039 (11) & 0.034 (16) & 0.074 (6) & 0.032 (19) & 0.036 (14) & 0.035 (15) & 0.039 (10) & 0.038 (12) & 0.049 (7) & 0.040 (9) & 0.042 (8) & 0.142 (4) & 0.033 (18) & 0.038 (13) & 0.077 (5) & 0.033 (17) & 0.164 (2) & 0.158 (3) & \textbf{0.164 (1)} \\
Annthyroid & 0.104 (12) & 0.113 (10) & 0.129 (8) & 0.139 (6) & 0.135 (7) & 0.129 (9) & 0.083 (17) & 0.090 (15) & 0.109 (11) & 0.066 (19) & 0.068 (18) & 0.093 (14) & 0.102 (13) & 0.171 (4) & 0.083 (16) & 0.155 (5) & 0.197 (3) & 0.218 (2) & \textbf{0.223 (1)} \\
Arrhythmia & 0.675 (14) & 0.765 (2) & 0.755 (6) & 0.751 (7) & 0.748 (9) & 0.739 (12) & 0.669 (15) & 0.637 (16) & 0.501 (19) & 0.747 (10) & 0.737 (13) & 0.750 (8) & 0.607 (17) & 0.591 (18) & 0.746 (11) & \textbf{0.768 (1)} & 0.761 (4) & 0.764 (3) & 0.757 (5) \\
Cardiotocography & 0.412 (8) & 0.437 (6) & 0.302 (16) & 0.473 (3) & 0.408 (9) & 0.398 (10) & 0.264 (18) & 0.362 (13) & 0.308 (15) & \textbf{0.531 (1)} & 0.520 (2) & 0.222 (19) & 0.270 (17) & 0.315 (14) & 0.392 (11) & 0.442 (5) & 0.429 (7) & 0.444 (4) & 0.384 (12) \\
Glass & 0.134 (11) & 0.153 (8) & 0.092 (19) & 0.197 (5) & 0.120 (14) & 0.136 (10) & 0.192 (6) & 0.121 (12) & 0.107 (17) & 0.117 (16) & 0.118 (15) & 0.172 (7) & 0.144 (9) & 0.120 (13) & 0.213 (2) & \textbf{0.253 (1)} & 0.209 (4) & 0.212 (3) & 0.102 (18) \\
HeartDisease & \textbf{0.598 (1)} & 0.541 (7) & 0.574 (2) & 0.525 (11) & 0.571 (3) & 0.566 (4) & 0.458 (16) & 0.441 (18) & 0.546 (6) & 0.536 (9) & 0.452 (17) & 0.514 (13) & 0.482 (14) & 0.420 (19) & 0.480 (15) & 0.523 (12) & 0.538 (8) & 0.532 (10) & 0.562 (5) \\
InternetAds & 0.331 (15) & 0.527 (5) & 0.366 (14) & 0.455 (10) & 0.482 (9) & 0.405 (11) & 0.296 (16) & 0.277 (18) & 0.202 (19) & 0.525 (6) & 0.503 (8) & 0.371 (13) & 0.295 (17) & 0.373 (12) & 0.576 (2) & 0.525 (7) & 0.534 (4) & 0.536 (3) & \textbf{0.579 (1)} \\
PageBlocks & 0.367 (16) & 0.465 (11) & 0.531 (5) & 0.409 (13) & 0.385 (15) & 0.390 (14) & 0.436 (12) & 0.262 (17) & 0.227 (18) & 0.551 (4) & 0.483 (7) & 0.192 (19) & 0.568 (2) & \textbf{0.575 (1)} & 0.564 (3) & 0.467 (9) & 0.480 (8) & 0.466 (10) & 0.494 (6) \\
PenDigits & 0.006 (13) & 0.005 (15) & 0.019 (5) & 0.006 (14) & 0.007 (11) & 0.007 (10) & 0.020 (4) & 0.014 (7) & \textbf{0.068 (1)} & 0.002 (19) & 0.002 (18) & 0.038 (2) & 0.013 (8) & 0.009 (9) & 0.015 (6) & 0.005 (17) & 0.007 (12) & 0.005 (16) & 0.021 (3) \\
Pima & 0.466 (11) & 0.516 (2) & 0.514 (3) & 0.436 (16) & 0.500 (6) & 0.488 (9) & 0.451 (13) & 0.409 (19) & 0.451 (12) & 0.410 (18) & 0.422 (17) & \textbf{0.527 (1)} & 0.438 (14) & 0.437 (15) & 0.474 (10) & 0.499 (7) & 0.509 (4) & 0.507 (5) & 0.491 (8) \\
Shuttle & 0.129 (7) & 0.069 (16) & 0.355 (4) & 0.090 (11) & 0.117 (8) & 0.114 (9) & 0.040 (17) & 0.175 (5) & 0.388 (3) & 0.022 (19) & 0.025 (18) & 0.160 (6) & \textbf{0.512 (1)} & 0.112 (10) & 0.084 (13) & 0.071 (15) & 0.090 (12) & 0.079 (14) & 0.396 (2) \\
SpamBase & 0.476 (9) & 0.480 (7) & 0.355 (18) & 0.533 (4) & 0.483 (6) & 0.439 (10) & 0.399 (14) & 0.349 (19) & 0.367 (17) & 0.404 (13) & 0.421 (12) & 0.377 (16) & 0.392 (15) & 0.431 (11) & 0.494 (5) & 0.479 (8) & 0.559 (2) & 0.557 (3) & \textbf{0.594 (1)} \\
Stamps & 0.249 (11) & 0.307 (8) & 0.333 (4) & 0.333 (5) & 0.333 (3) & 0.256 (10) & 0.196 (13) & 0.113 (19) & 0.164 (16) & 0.123 (18) & 0.146 (17) & 0.174 (15) & 0.213 (12) & 0.189 (14) & 0.334 (2) & \textbf{0.345 (1)} & 0.309 (7) & 0.327 (6) & 0.269 (9) \\
WBC & 0.556 (12) & 0.882 (3) & 0.875 (5) & 0.827 (9) & 0.864 (8) & 0.604 (11) & 0.274 (18) & 0.537 (13) & 0.368 (17) & 0.394 (16) & 0.501 (14) & 0.750 (10) & 0.111 (19) & 0.400 (15) & \textbf{0.895 (1)} & 0.877 (4) & 0.874 (6) & 0.885 (2) & 0.865 (7) \\
WDBC & 0.647 (11) & 0.647 (10) & 0.691 (7) & 0.716 (4) & 0.697 (5) & 0.684 (8) & 0.092 (19) & 0.551 (13) & 0.316 (16) & 0.630 (12) & 0.501 (15) & 0.512 (14) & 0.197 (18) & 0.275 (17) & \textbf{0.800 (1)} & 0.678 (9) & 0.760 (3) & 0.694 (6) & 0.768 (2) \\
WPBC & 0.233 (7) & 0.231 (9) & 0.232 (8) & 0.239 (5) & 0.231 (10) & 0.230 (11) & 0.243 (4) & 0.206 (19) & 0.266 (2) & 0.229 (12) & 0.207 (18) & 0.227 (14) & \textbf{0.275 (1)} & 0.216 (17) & 0.265 (3) & 0.227 (15) & 0.229 (13) & 0.225 (16) & 0.235 (6) \\
Waveform & 0.078 (8) & 0.061 (12) & 0.131 (4) & 0.055 (16) & 0.066 (9) & 0.081 (7) & 0.029 (19) & 0.033 (18) & 0.056 (15) & 0.058 (13) & 0.066 (10) & 0.141 (3) & 0.035 (17) & 0.063 (11) & 0.168 (2) & 0.056 (14) & 0.115 (6) & 0.118 (5) & \textbf{0.190 (1)} \\
Wilt & 0.048 (13) & 0.045 (17) & 0.053 (8) & 0.041 (19) & 0.043 (18) & 0.055 (6) & \textbf{0.283 (1)} & 0.089 (3) & 0.046 (14) & 0.054 (7) & 0.052 (9) & 0.083 (4) & 0.147 (2) & 0.049 (10) & 0.079 (5) & 0.045 (16) & 0.048 (12) & 0.046 (15) & 0.048 (11) \\
annthyroid & 0.195 (13) & 0.314 (7) & 0.204 (12) & 0.366 (5) & 0.285 (8) & 0.247 (10) & 0.161 (15) & 0.128 (17) & 0.120 (18) & 0.135 (16) & 0.114 (19) & 0.173 (14) & \textbf{0.655 (1)} & 0.243 (11) & 0.257 (9) & 0.336 (6) & 0.452 (3) & 0.395 (4) & 0.531 (2) \\
arrhythmia & 0.424 (14) & 0.479 (4) & 0.464 (7) & \textbf{0.502 (1)} & 0.445 (11) & 0.431 (13) & 0.313 (15) & 0.257 (19) & 0.309 (16) & 0.460 (8) & 0.450 (10) & 0.442 (12) & 0.291 (17) & 0.267 (18) & 0.459 (9) & 0.482 (2) & 0.474 (6) & 0.480 (3) & 0.475 (5) \\
breastw & 0.895 (9) & 0.969 (4) & 0.392 (18) & 0.967 (5) & 0.979 (2) & 0.941 (8) & 0.883 (10) & 0.782 (12) & 0.320 (19) & 0.686 (15) & 0.728 (14) & 0.748 (13) & 0.786 (11) & 0.569 (17) & 0.588 (16) & \textbf{0.979 (1)} & 0.955 (6) & 0.955 (7) & 0.972 (3) \\
glass & 0.076 (16) & 0.093 (10) & 0.083 (13) & 0.116 (7) & 0.085 (12) & 0.080 (14) & 0.195 (2) & 0.052 (18) & 0.041 (19) & \textbf{0.205 (1)} & 0.154 (5) & 0.171 (4) & 0.172 (3) & 0.078 (15) & 0.125 (6) & 0.116 (8) & 0.092 (11) & 0.104 (9) & 0.058 (17) \\
ionosphere & 0.636 (17) & 0.809 (6) & 0.799 (9) & 0.758 (13) & 0.803 (8) & 0.745 (14) & \textbf{0.949 (1)} & 0.599 (18) & 0.565 (19) & 0.735 (16) & 0.742 (15) & 0.914 (3) & 0.916 (2) & 0.809 (7) & 0.790 (11) & 0.910 (4) & 0.793 (10) & 0.813 (5) & 0.759 (12) \\
letter & 0.140 (7) & 0.087 (15) & 0.244 (4) & 0.088 (12) & 0.130 (8) & 0.220 (5) & 0.311 (2) & 0.094 (10) & 0.081 (19) & 0.100 (9) & 0.085 (16) & \textbf{0.383 (1)} & 0.269 (3) & 0.088 (14) & 0.209 (6) & 0.091 (11) & 0.085 (17) & 0.088 (13) & 0.082 (18) \\
lympho & 0.782 (10) & 0.944 (3) & 0.857 (6) & 0.976 (2) & 0.877 (5) & 0.822 (8) & 0.151 (18) & 0.519 (14) & 0.056 (19) & 0.897 (4) & 0.808 (9) & 0.640 (12) & 0.323 (16) & 0.403 (15) & 0.173 (17) & \textbf{1.000 (1)} & 0.720 (11) & 0.852 (7) & 0.556 (13) \\
mammography & 0.184 (9) & 0.221 (6) & 0.121 (12) & 0.221 (7) & 0.247 (3) & 0.198 (8) & 0.171 (11) & 0.100 (15) & 0.070 (19) & 0.073 (18) & 0.097 (16) & 0.105 (13) & 0.176 (10) & 0.102 (14) & 0.082 (17) & 0.238 (4) & 0.269 (2) & 0.235 (5) & \textbf{0.284 (1)} \\
mnist & 0.246 (16) & 0.265 (13) & 0.379 (5) & 0.262 (14) & 0.278 (12) & 0.290 (9) & 0.393 (3) & 0.225 (17) & 0.207 (18) & 0.392 (4) & 0.333 (8) & 0.350 (7) & 0.412 (2) & \textbf{0.442 (1)} & 0.371 (6) & 0.246 (15) & 0.282 (11) & 0.282 (10) & 0.160 (19) \\
musk & 0.548 (13) & \textbf{1.000 (1)} & 0.090 (17) & 1.000 (2) & 1.000 (3) & 0.882 (10) & 0.450 (15) & 0.856 (11) & 0.043 (18) & 0.998 (8) & 0.440 (16) & 0.025 (19) & 0.517 (14) & 0.688 (12) & 0.994 (9) & 1.000 (4) & 1.000 (5) & 1.000 (6) & 1.000 (7) \\
optdigits & 0.038 (12) & 0.051 (6) & 0.021 (18) & 0.043 (10) & 0.054 (5) & 0.050 (7) & 0.023 (16) & 0.018 (19) & 0.032 (13) & 0.070 (3) & 0.043 (11) & 0.021 (17) & 0.029 (14) & 0.028 (15) & 0.072 (2) & 0.055 (4) & 0.050 (8) & 0.048 (9) & \textbf{0.114 (1)} \\
pendigits & 0.195 (10) & 0.279 (3) & 0.044 (16) & 0.255 (7) & 0.260 (6) & 0.214 (9) & 0.035 (18) & 0.083 (13) & 0.036 (17) & 0.182 (11) & 0.228 (8) & 0.033 (19) & 0.048 (15) & 0.173 (12) & 0.066 (14) & 0.266 (5) & 0.289 (2) & 0.272 (4) & \textbf{0.359 (1)} \\
pima & 0.477 (11) & 0.500 (6) & 0.493 (7) & 0.465 (12) & 0.503 (5) & \textbf{0.521 (1)} & 0.405 (15) & 0.431 (13) & 0.398 (16) & 0.368 (18) & 0.382 (17) & 0.504 (4) & 0.431 (14) & 0.349 (19) & 0.490 (8) & 0.505 (3) & 0.480 (10) & 0.487 (9) & 0.516 (2) \\
satellite & 0.604 (9) & 0.660 (4) & 0.397 (18) & 0.660 (5) & 0.680 (2) & 0.613 (8) & 0.508 (15) & 0.574 (12) & 0.479 (16) & 0.527 (14) & 0.587 (10) & 0.332 (19) & 0.580 (11) & 0.540 (13) & 0.415 (17) & 0.658 (6) & 0.658 (7) & 0.671 (3) & \textbf{0.698 (1)} \\
satimage-2 & 0.709 (11) & 0.926 (4) & 0.142 (16) & 0.915 (7) & 0.944 (2) & 0.754 (10) & 0.100 (17) & 0.394 (12) & 0.063 (18) & \textbf{0.961 (1)} & 0.889 (9) & 0.033 (19) & 0.142 (15) & 0.380 (14) & 0.393 (13) & 0.921 (6) & 0.927 (3) & 0.924 (5) & 0.895 (8) \\
speech & 0.027 (3) & 0.018 (18) & 0.020 (11) & 0.027 (5) & 0.021 (10) & 0.028 (2) & 0.019 (13) & 0.023 (9) & 0.027 (4) & 0.019 (14) & 0.019 (12) & 0.024 (8) & 0.018 (17) & \textbf{0.038 (1)} & 0.019 (15) & 0.018 (16) & 0.026 (6) & 0.026 (7) & 0.017 (19) \\
thyroid & 0.464 (9) & 0.557 (7) & 0.335 (13) & 0.380 (12) & 0.497 (8) & 0.434 (10) & 0.309 (14) & 0.146 (16) & 0.049 (19) & 0.176 (15) & 0.118 (18) & 0.120 (17) & 0.729 (2) & 0.565 (6) & 0.410 (11) & 0.612 (4) & 0.651 (3) & 0.604 (5) & \textbf{0.746 (1)} \\
vertebral & 0.095 (12) & 0.096 (11) & 0.088 (17) & 0.092 (15) & 0.087 (18) & 0.090 (16) & 0.116 (5) & 0.144 (4) & 0.103 (7) & 0.156 (3) & 0.174 (2) & 0.099 (8) & 0.115 (6) & \textbf{0.193 (1)} & 0.000 (19) & 0.098 (10) & 0.094 (13) & 0.093 (14) & 0.099 (9) \\
vowels & 0.178 (9) & 0.138 (13) & 0.385 (4) & 0.106 (15) & 0.233 (7) & 0.195 (8) & 0.458 (3) & 0.067 (17) & 0.057 (19) & 0.140 (12) & 0.063 (18) & \textbf{0.615 (1)} & 0.362 (6) & 0.089 (16) & 0.369 (5) & 0.127 (14) & 0.155 (11) & 0.165 (10) & 0.491 (2) \\
wbc & 0.612 (6) & 0.608 (7) & 0.650 (3) & 0.582 (11) & 0.597 (8) & 0.616 (5) & 0.241 (18) & 0.415 (16) & 0.447 (14) & 0.596 (9) & 0.417 (15) & 0.378 (17) & 0.203 (19) & 0.472 (13) & 0.542 (12) & 0.591 (10) & 0.639 (4) & \textbf{0.668 (1)} & 0.663 (2) \\
wine & 0.246 (6) & 0.213 (12) & 0.290 (2) & 0.234 (8) & 0.255 (5) & 0.274 (3) & 0.093 (17) & 0.112 (15) & 0.081 (18) & 0.220 (9) & 0.108 (16) & 0.239 (7) & 0.256 (4) & 0.123 (14) & 0.000 (19) & 0.211 (13) & 0.215 (10) & 0.215 (11) & \textbf{0.366 (1)} \\
\midrule
\textbf{Average} & 0.342 (10.56) & 0.398 (8.31) & 0.330 (9.49) & 0.392 (9.03) & 0.397 (8.00) & 0.369 (9.08) & 0.276 (12.18) & 0.275 (14.18) & 0.208 (14.62) & 0.347 (11.00) & 0.314 (12.82) & 0.302 (10.67) & 0.314 (10.67) & 0.293 (12.08) & 0.348 (9.08) & 0.408 (7.95) & 0.413 (7.08) & 0.414 (7.00) & \textbf{0.435 (6.23)} \\
\textbf{Std Dev} & 0.246 & 0.308 & 0.248 & 0.302 & 0.301 & 0.268 & 0.223 & 0.227 & 0.171 & 0.285 & 0.255 & 0.242 & 0.231 & 0.209 & 0.267 & 0.313 & 0.294 & 0.297 & 0.285 \\
\bottomrule
\end{tabular}
}
\end{table*}

\subsection{Pool Size Analysis}
\label{app:sec:pool_size_analysis}

We study the performance and robustness of \framework by varying the size of the candidate model pool from 10 to 297 detectors. For each pool size, we randomly sample the specified number of models and apply the full \framework selection procedure. To account for variability introduced by sub-pool sampling, experiments are repeated with 10 random seeds and results are averaged.

\begin{figure}[t]
\centering
\includegraphics[width=0.7\linewidth]{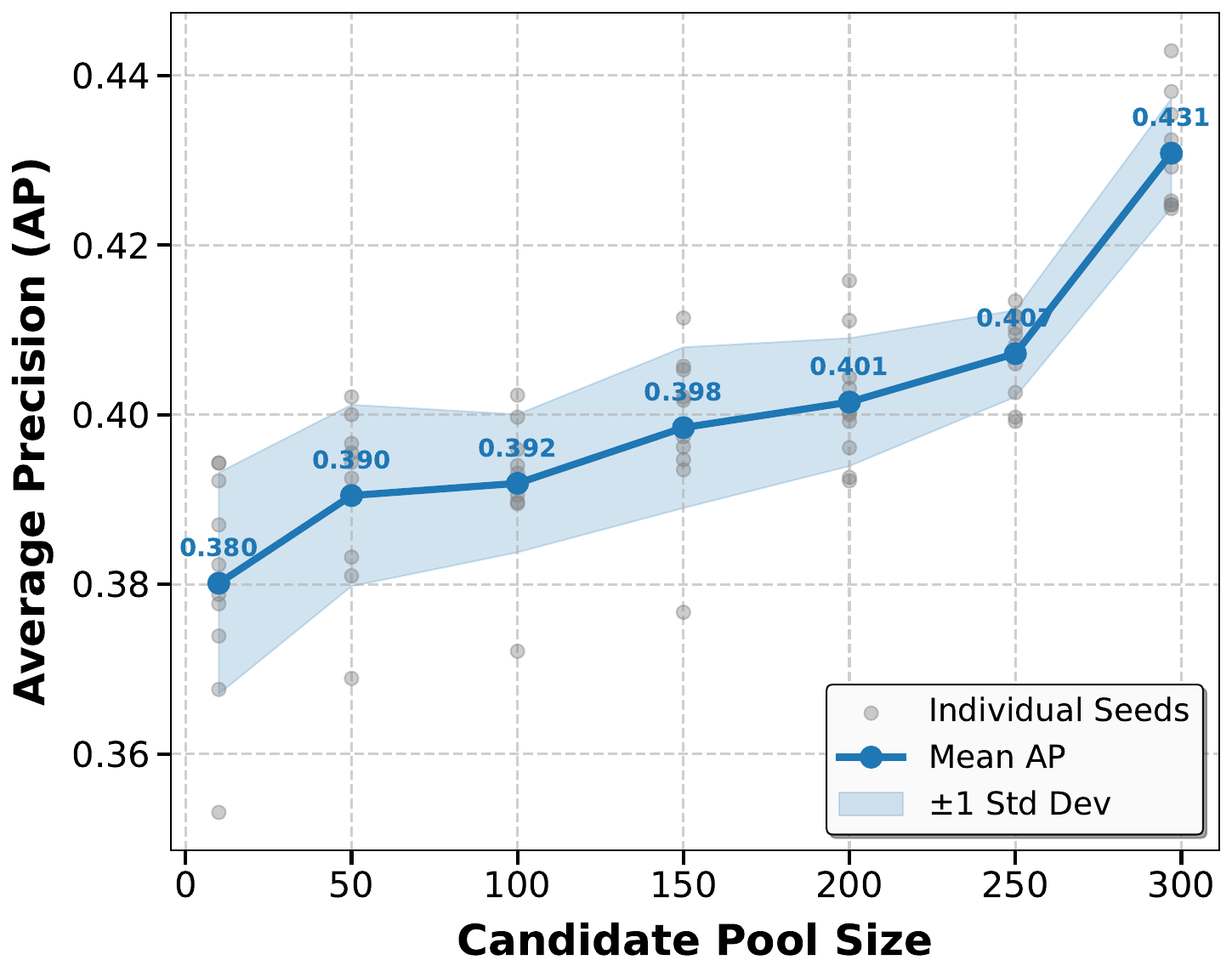}
\caption{Impact of candidate pool size on \framework performance. Results are averaged over 10 random seeds for each pool size. The shaded area represents $\pm 1$ standard deviation.}
\label{fig:pool_size_analysis}
\end{figure}

Figure~\ref{fig:pool_size_analysis} shows the relationship between pool size and final ensemble performance measured by AP. The blue curve denotes the average AP across the 10 runs, and the shaded region indicates $\pm 1$ standard deviation.

Performance improves steadily as the pool size increases, indicating that \framework effectively leverages a richer set of candidate models. Gains begin to plateau around 250 models, suggesting diminishing returns beyond this point. This trend indicates that the full pool of 297 models is sufficient to capture the diversity needed for robust anomaly detection across the benchmark datasets.

To separate algorithm diversity from hyperparameter selection, we further run a single-family pool analysis. In this setting, \framework may still select and combine multiple hyperparameter variants, but all candidates come from the same algorithmic family. Table~\ref{tab:single_family_pool} shows that performance falls substantially for IForest-only and LOF-only pools, while the average ensemble size remains close to the full setting. Thus, the loss is mainly due to missing cross-family candidates rather than an inability to form ensembles, indicating that algorithm diversity is the primary driver and hyperparameter selection provides complementary gains.

\begin{table}[t]
\scriptsize
\caption{Single-family pool analysis. Restricted pools remove cross-family diversity.}
\label{tab:single_family_pool}
\centering
\setlength{\tabcolsep}{3pt}
\resizebox{\columnwidth}{!}{%
\begin{tabular}{lccc}
\toprule
\textbf{Candidate Pool} & \textbf{AP} $\uparrow$ & \textbf{Rank} $\downarrow$ & \textbf{Ens. Size} \\
\midrule
IForest-only (81) & $0.3552 \pm 0.0015$ & N/A & 2.0 \\
LOF-only (36) & $0.3369 \pm 0.0028$ & N/A & 2.0 \\
Full 8-family (297) & $\mathbf{0.4308 \pm 0.0064}$ & $\mathbf{59.3 \pm 6.9}$ & \textbf{2.2} \\
\bottomrule
\end{tabular}
}
\end{table}

\subsection{Statistical Significance Analysis}
\label{app:sec:statistical_significance}

As discussed in Sec.~\ref{subsec:statistical_significance}, we use paired Wilcoxon signed-rank tests to compare \framework against each baseline across the 39 benchmark datasets. The test is non-parametric and is therefore suitable for paired comparisons across heterogeneous datasets without assuming normally distributed AP differences. We use a one-sided alternative hypothesis that \framework achieves higher AP than the baseline and set $\alpha=0.05$.
Table~\ref{tab:statistical_significance} reports the corresponding $p$-values, mean AP differences, and dataset-level win rates.

\begin{table}[t]
\fontsize{8}{8}\selectfont
\caption{Statistical significance testing: Paired Wilcoxon signed-rank test comparing \texttt{MetaEns} against baselines across 39 datasets (one-sided test, $\alpha=0.05$).}
\label{tab:statistical_significance}
\centering
\begin{tabular}{ll|c|cc}
\toprule
Ours & Baseline & $p$-value & $\Delta$AP & Win\% \\
\midrule
\framework & \texttt{Random Selection} & 0.0000 & +0.0939 & 79.5\% \\
\framework & \texttt{IForest} & 0.0151 & +0.0378 & 64.1\% \\
\framework & \texttt{LOF} & 0.0022 & +0.1052 & 69.2\% \\
\framework & \texttt{Global Best} & 0.0050 & +0.0438 & 71.8\% \\
\framework & \texttt{Mega Ensemble} & 0.0096 & +0.0384 & 66.7\% \\
\framework & \texttt{Random Ensemble} & 0.0002 & +0.0662 & 74.4\% \\
\framework & \texttt{RDA} & 0.0001 & +0.1599 & 76.9\% \\
\framework & \texttt{DAGMM} & 0.0000 & +0.1605 & 84.6\% \\
\framework & \texttt{DeepSVDD}  & 0.0000 & +0.2272 & 84.6\% \\
\framework & \texttt{ROBOD} & 0.0000 & +0.1173 & 79.5\% \\
\framework & \texttt{LUNAR} & 0.0004 & +0.1284 & 71.8\% \\
\framework & \texttt{DTE-C} & 0.0010 & +0.1164 & 71.8\% \\
\framework & \texttt{TCCM} & 0.0000 & +0.1379 & 76.9\% \\
\framework & \texttt{RandNet} & 0.0013 & +0.0884 & 71.8\% \\
\framework & \texttt{LSCP} & 0.0047 & +0.0876 & 66.7\% \\
\framework & \texttt{MetaOD} & 0.0390 & +0.0275 & 61.5\% \\
\framework & \texttt{ELECT-1} & 0.0314 & +0.0222 & 66.7\% \\
\framework & \texttt{ELECT-10} & 0.0386 & +0.0213 & 61.5\% \\
\bottomrule
\end{tabular}
\end{table}

\subsection{Modality Transfer Experiments}
\label{app:subsec:modality_transfer}

We further evaluate whether \framework can be used as a plug-and-play selector outside tabular anomaly detection. Since the state representation depends only on detector-score vectors, no architectural change is needed for image or text settings. We use ADBench extracted features: \texttt{CV\_by\_ResNet18}
(512-dimensional ResNet-18 features)  or all 15 MVTec-AD image categories
(\texttt{bottle}, \texttt{cable}, \texttt{capsule}, \texttt{carpet},
\texttt{grid}, \texttt{hazelnut}, \texttt{leather}, \texttt{metal\_nut},
\texttt{pill}, \texttt{screw}, \texttt{tile}, \texttt{toothbrush},
\texttt{transistor}, \texttt{wood}, and \texttt{zipper}) and
\texttt{NLP\_by\_BERT} (768-dimensional BERT features) for 5 text datasets.
The compared methods are \framework, random singleton selection, mega ensemble,
\texttt{ELECT-1}, \texttt{ELECT-10}, \texttt{ROBOD}, \texttt{LUNAR},
\texttt{DTE-C}, and \texttt{TCCM}, all under the same unsupervised evaluation
protocol.

\begin{table}[t]
\scriptsize
\caption{Modality-transfer results on 20 non-tabular datasets.}
\label{tab:modality_transfer_summary}
\centering
\setlength{\tabcolsep}{3pt}
\resizebox{\columnwidth}{!}{%
\begin{tabular}{lcccc}
\toprule
\textbf{Setting} & \textbf{\#Data} & \textbf{\framework AP} $\uparrow$ & \textbf{Best Base. AP} $\uparrow$ & \textbf{Gap} \\
\midrule
Overall & 20 & $\mathbf{0.4914 \pm 0.0007}$ & \texttt{LUNAR}: $0.4717 \pm 0.0045$ & $+0.0197$ \\
Image (MVTec-AD) & 15 & $\mathbf{0.6230 \pm 0.0009}$ & \texttt{ELECT-1}: $0.5973 \pm 0.0035$ & $+0.0257$ \\
Text & 5 & $0.0966$ & \texttt{LUNAR}: $0.0977 \pm 0.0021$ & $-0.0012$ \\
\bottomrule
\end{tabular}
}
\end{table}

Table~\ref{tab:modality_transfer_summary} reports the headline AP results. Across all 20 datasets, \framework reaches $0.4914\pm0.0007$ AP, outperforming the strongest baseline, \texttt{LUNAR} ($0.4717\pm0.0045$). On MVTec-AD image data, \framework obtains $0.6230\pm0.0009$ AP, improving over \texttt{ELECT-1} ($0.5973\pm0.0035$). Table~\ref{tab:modality_transfer_stats} further reports win/loss statistics and one-sided Wilcoxon tests, confirming significant gains against all compared baselines over the 20 non-tabular datasets.

\begin{table}[t]
\fontsize{8}{8}\selectfont
\caption{Dataset-level win/loss and one-sided Wilcoxon signed-rank tests for modality-transfer experiments over 20 non-tabular datasets.}
\label{tab:modality_transfer_stats}
\centering
\setlength{\tabcolsep}{5pt}
\begin{tabular}{lcc}
\toprule
\textbf{Baseline} & \textbf{Win/Loss} & \textbf{$p$-value} \\
\midrule
\texttt{ELECT-1} & 14 / 6 & 0.0148 \\
\texttt{ELECT-10} & 17 / 3 & 0.0047 \\
\texttt{LUNAR} & 15 / 5 & 0.0266 \\
\texttt{ROBOD} & 18 / 2 & $1.61\times 10^{-4}$ \\
\texttt{DTE-C} & 18 / 2 & $6.7\times 10^{-5}$ \\
\texttt{TCCM} & 18 / 2 & $1.05\times 10^{-4}$ \\
\texttt{Mega Ensemble} & 18 / 2 & $5.1\times 10^{-4}$ \\
\texttt{Random Singleton} & 19 / 1 & $6.7\times 10^{-5}$ \\
\bottomrule
\end{tabular}
\end{table}

\subsection{Model Diversity Visualization}
\label{app:sec:mode_diversity}

\begin{table}[t]
\fontsize{8}{8}\selectfont
\caption{Quantitative diversity of selected ensembles on the 39 tabular datasets. Distinct families count unique base algorithms; pairwise Jaccard is the mean top-10\% overlap between selected detectors.}
\label{tab:diversity_quantitative}
\centering
\setlength{\tabcolsep}{4pt}
\begin{tabular}{lcc}
\toprule
\textbf{Method} & \textbf{Distinct Families} $\uparrow$ & \textbf{Pairwise Jaccard} $\downarrow$ \\
\midrule
\texttt{ELECT} (Top-10) & 2.1 & 0.68 \\
\texttt{Random Ensemble} & 1.0 & --- \\
\texttt{Mega Ensemble} & 8.0 & 0.44 \\
\framework{} (Ours) & \textbf{2.2} & \textbf{0.36} \\
\bottomrule
\end{tabular}
\end{table}

As a qualitative complement to Table~\ref{tab:diversity_quantitative}, we project the 297-model candidate pool into a two-dimensional space using \texttt{t-SNE}~\cite{DBLP:journals/JMLR/MaatenH08} based on prediction-correlation distance. The embedding shows that detectors naturally cluster by algorithmic family, since variants within the same family tend to produce similar outlier rankings.

Figure~\ref{fig:diversity_tsne} visualizes four representative datasets (\textbf{Speech}, \textbf{WBC}, \textbf{Waveform}, and \textbf{Shuttle}). \texttt{ELECT-10} selections often concentrate within a single family cluster, whereas \framework selects models spanning multiple clusters. This supports the quantitative finding that \framework constructs compact but complementary ensembles rather than simply adding many redundant high-ranked detectors.

\subsection{Effect of Expanded Model Pools}
\label{app:subsec:expanded_model_pool}

To evaluate \framework's scalability and robustness to model pool composition, we conduct additional experiments with an expanded pool of 310 models that incorporates neural network variants alongside the original 297 classical detectors.

\paragraph{Expanded Pool Composition.}

The expanded pool adds 13 neural network variants spanning 5 deep learning families: \texttt{AutoEncoder}~\cite{DBLP:conf/sdm/ChenSAT17} (4 architectural variants),  \texttt{Variational AutoEncoder}~\cite{DBLP:conf/www/XuCZLBLLZPFCWQ18} (3 variants), \texttt{SO\_GAAL}~\cite{DBLP:journals/tkde/LiuLZJSWH20} (2 configurations), \texttt{MO\_GAAL}~\cite{DBLP:journals/tkde/LiuLZJSWH20} (2 configurations), and \texttt{DeepSVDD}~\cite{DBLP:conf/icml/RuffGDSVBMK18} (2 variants). These models represent diverse deep learning paradigms, including reconstruction-based methods, adversarial approaches, and deep one-class classification. Table~\ref{tab:model_pool_nn} provides the complete specification of the 310-model expanded pool.

\begin{table*}[t]
\centering
\caption{Expanded candidate model pool specification with neural network variants ($M = 310$ total models).}
\label{tab:model_pool_nn}
\centering
\small
\begin{tabular}{llp{7cm}}
\toprule
\textbf{Family} & \textbf{Count} & \textbf{Hyperparameter Grid} \\
\midrule
\texttt{kNN}~\cite{DBLP:conf/sdm/Chehreghani16} & 36 & \makecell{method $\in \{\text{largest}, \text{mean}, \text{median}\}$;\\ $k \in \{1, 5, 10, 15, 20, 25, 50, 60, 70, 80, 90, 100\}$} \\
\midrule
\texttt{LOF}~\cite{DBLP:conf/sigmod/BreunigKNS00} & 36 & \makecell{metric $\in \{\text{euclidean}, \text{manhattan}, \text{minkowski}\}$; \\$k \in \{1, 5, 10, 15, 20, 25, 50, 60, 70, 80, 90, 100\}$} \\
\midrule
\texttt{IForest}~\cite{DBLP:conf/icdm/LiuTZ08} & 81 & \makecell{$n_{\text{estimators}} \in \{10, 20, 30, 40, 50, 75, 100, 150, 200\}$; \\max\_samples $\in \{0.1, 0.2, \ldots, 0.9\}$} \\
\midrule
\texttt{HBOS}~\cite{goldstein2012histogram} & 40 & \makecell{$n_{\text{bins}} \in \{5, 10, 20, 30, 40, 50, 75, 100\}$; \\tolerance $\in \{0.1, 0.2, 0.3, 0.4, 0.5\}$} \\
\midrule
\texttt{OCSVM}~\cite{DBLP:conf/nips/ScholkopfWSSP99} & 36 & \makecell{kernel $\in \{\text{linear}, \text{poly}, \text{rbf}, \text{sigmoid}\}$; \\$\nu \in \{0.1, 0.2, \ldots, 0.9\}$} \\
\midrule
\texttt{LODA}~\cite{DBLP:journals/ml/Pevny16} & 54 & \makecell{$n_{\text{bins}} \in \{5, 10, 15, 20, 25, 30\}$; \\$n_{\text{cuts}} \in \{10, 20, 30, 40, 50, 75, 100, 150, 200\}$} \\
\midrule
\texttt{ABOD}~\cite{DBLP:conf/kdd/KriegelSZ08} & 7 & $n_{\text{neighbors}} \in \{3, 5, 10, 15, 20, 25, 50\}$ \\
\midrule
\texttt{COF}~\cite{DBLP:conf/pakdd/TangCFC02} & 7 & $n_{\text{neighbors}} \in \{3, 5, 10, 15, 20, 25, 50\}$ \\
\midrule
\multicolumn{3}{c}{\textbf{Neural Network Extensions}} \\
\midrule
\texttt{Autoencoder}~\cite{DBLP:conf/sdm/ChenSAT17} & 4 & Architecture variants: \texttt{[64,32,32,64]}, \texttt{[128,64,64,128]}, \texttt{[32,16,16,32]}, \texttt{[64,32,16,16,32,64]} \\
\midrule
\texttt{Variational Autoencoder}~\cite{DBLP:conf/www/XuCZLBLLZPFCWQ18} & 3 & Encoder-decoder pairs: \texttt{([64,32],[32,64])}, \texttt{([128,64],[64,128])}, \texttt{([32,16],[16,32])} \\
\midrule
\texttt{SO\_GAAL}~\cite{DBLP:journals/tkde/LiuLZJSWH20} & 2 & Single-objective GAN: \texttt{stop\_epochs}=20; lr $\in \{0.0005, 0.001\}$ \\
\midrule
\texttt{MO\_GAAL}~\cite{DBLP:journals/tkde/LiuLZJSWH20} & 2 & Multi-objective GAN: \texttt{stop\_epochs}=20; $k \in \{5, 20\}$ \\
\midrule
\texttt{DeepSVDD}~\cite{DBLP:conf/icml/RuffGDSVBMK18} & 2 & Deep SVDD: pre-training $\in \{\text{True}, \text{False}\}$ \\
\midrule
\textbf{Classical Total} & \textbf{297} & Traditional anomaly detection methods \\
\textbf{Neural Network Total} & \textbf{13} & Deep learning-based detection methods \\
\textbf{Grand Total} & \textbf{310} & Complete expanded model pool \\
\bottomrule
\end{tabular}
\end{table*}

\paragraph{Overall Performance Comparison.}
Table~\ref{tab:nn_pool_scalability} presents the scalability analysis results comparing \framework performance on the original (297 models) versus expanded (310 models) pools. \framework demonstrates robust performance across both pool configurations, maintaining competitive efficacy with only a modest decrease in Average Precision ($0.0186$) and an increase in median rank ($13.0$).


\begin{table*}[ht]
\centering
\caption{\framework scalability analysis: Performance on original (297) vs. expanded (310) model pools. Results demonstrate method robustness across different pool compositions.}
\label{tab:nn_pool_scalability}
\small
\begin{tabular}{@{}llllll@{}}
\toprule
\textbf{Pool Configuration} & \textbf{Size} & \textbf{AP (Mean $\pm$ Std)} $\uparrow$ & \textbf{Rank} $\downarrow$ & \textbf{ROC-AUC} $\uparrow$ & \textbf{Ens Size} \\
\midrule
Classical Models Only & 297 & 0.4308 $\pm$ 0.0064 & 59.3 $\pm$ 6.9610 & 0.7867 $\pm$ 0.0045 & 2.2 \\
Classical + Neural Networks & 310 & 0.4122 $\pm$ 0.0068 & 72.3 $\pm$ 11.4018 & 0.7823 $\pm$ 0.0044 & 2.3 \\
\bottomrule
\end{tabular}%
\end{table*}

\paragraph{Detailed Dataset-by-Dataset Results.}
Table~\ref{tab:detailed_performance_nn_pool} presents the complete performance comparison of all methods on the expanded 310-model pool. The table shows Average Precision (AP) values and model ranks (1--310, where lower ranks indicate better performance) for each method across all 39 benchmark datasets, with best results for each dataset highlighted in bold.

\begin{table*}[t]
\centering
\caption{Detailed Performance Comparison Across 39 Benchmark Datasets on Expanded Neural Network Pool (310 models). For each method, we report Average Precision (AP) and rank in parentheses (lower rank is better, 1--19, among compared methods). Best results in \textbf{bold}. Per-dataset values are from a single representative seed (seed=42). Abbreviations: RS = Random Selection, RE = Random Ensemble (k=3), ELECT-1 = ELECT (Top-1), ELECT-10 = ELECT (Top-10).}
\label{tab:detailed_performance_nn_pool}
\small
\setlength{\tabcolsep}{3pt}
\resizebox{\textwidth}{!}{
\begin{tabular}{l|llllllllllllllllll|l}
\toprule
\textbf{Dataset} & \textbf{RS} & \textbf{IForest} & \textbf{LOF} & \textbf{GB} & \textbf{ME} & \textbf{RE} & \textbf{RDA} & \textbf{DAGMM} & \textbf{DeepSVDD} & \textbf{RandNet} & \textbf{ROBOD} & \textbf{LUNAR} & \textbf{DTE-C} & \textbf{TCCM} & \textbf{LSCP} & \textbf{MetaOD} & \textbf{ELECT-1} & \textbf{ELECT-10} & \textbf{MetaEns} \\
\midrule
ALOI & 0.039 (10) & 0.034 (16) & 0.074 (6) & 0.032 (19) & 0.036 (14) & 0.035 (15) & 0.039 (11) & 0.038 (12) & 0.049 (7) & 0.040 (9) & 0.042 (8) & 0.142 (4) & 0.033 (18) & 0.038 (13) & 0.077 (5) & 0.033 (17) & \textbf{0.164 (1)} & 0.158 (3) & 0.164 (2) \\
Annthyroid & 0.104 (12) & 0.113 (10) & 0.129 (8) & 0.139 (6) & 0.136 (7) & 0.129 (9) & 0.083 (16) & 0.090 (15) & 0.109 (11) & 0.066 (19) & 0.068 (18) & 0.093 (14) & 0.102 (13) & 0.171 (4) & 0.083 (17) & 0.155 (5) & 0.197 (2) & \textbf{0.218 (1)} & 0.197 (3) \\
Arrhythmia & 0.675 (14) & 0.765 (2) & 0.755 (5) & 0.751 (6) & 0.746 (9) & 0.737 (11) & 0.669 (15) & 0.637 (16) & 0.501 (19) & 0.747 (8) & 0.737 (12) & 0.750 (7) & 0.607 (17) & 0.591 (18) & 0.746 (10) & \textbf{0.768 (1)} & 0.761 (4) & 0.764 (3) & 0.726 (13) \\
Cardiotocography & 0.412 (8) & 0.437 (6) & 0.302 (16) & 0.473 (3) & 0.410 (9) & 0.400 (10) & 0.264 (18) & 0.362 (13) & 0.308 (15) & \textbf{0.531 (1)} & 0.520 (2) & 0.222 (19) & 0.270 (17) & 0.315 (14) & 0.392 (11) & 0.442 (5) & 0.429 (7) & 0.444 (4) & 0.384 (12) \\
Glass & 0.134 (10) & 0.153 (8) & 0.092 (19) & 0.197 (5) & 0.119 (14) & 0.131 (11) & 0.192 (6) & 0.121 (12) & 0.107 (17) & 0.117 (16) & 0.118 (15) & 0.172 (7) & 0.144 (9) & 0.120 (13) & 0.213 (2) & \textbf{0.253 (1)} & 0.209 (4) & 0.212 (3) & 0.102 (18) \\
HeartDisease & \textbf{0.598 (1)} & 0.541 (7) & 0.574 (2) & 0.525 (11) & 0.567 (3) & 0.566 (4) & 0.458 (16) & 0.441 (18) & 0.546 (6) & 0.536 (9) & 0.452 (17) & 0.514 (13) & 0.482 (14) & 0.420 (19) & 0.480 (15) & 0.523 (12) & 0.538 (8) & 0.532 (10) & 0.562 (5) \\
InternetAds & 0.339 (15) & 0.527 (5) & 0.366 (14) & 0.455 (10) & 0.476 (9) & 0.390 (11) & 0.296 (16) & 0.277 (18) & 0.202 (19) & 0.525 (6) & 0.503 (8) & 0.371 (13) & 0.295 (17) & 0.373 (12) & 0.576 (2) & 0.525 (7) & 0.534 (4) & 0.536 (3) & \textbf{0.579 (1)} \\
PageBlocks & 0.325 (16) & 0.465 (11) & 0.531 (5) & 0.409 (13) & 0.393 (14) & 0.368 (15) & 0.436 (12) & 0.262 (17) & 0.227 (18) & 0.551 (4) & 0.483 (7) & 0.192 (19) & 0.568 (2) & \textbf{0.575 (1)} & 0.564 (3) & 0.467 (9) & 0.480 (8) & 0.466 (10) & 0.494 (6) \\
PenDigits & 0.006 (13) & 0.005 (15) & 0.019 (5) & 0.006 (14) & 0.007 (10) & 0.007 (11) & 0.020 (4) & 0.014 (7) & \textbf{0.068 (1)} & 0.002 (19) & 0.002 (18) & 0.038 (2) & 0.013 (8) & 0.009 (9) & 0.015 (6) & 0.005 (16) & 0.007 (12) & 0.005 (17) & 0.021 (3) \\
Pima & 0.466 (11) & 0.516 (2) & 0.514 (3) & 0.436 (16) & 0.497 (8) & 0.488 (9) & 0.451 (12) & 0.409 (19) & 0.451 (13) & 0.410 (18) & 0.422 (17) & \textbf{0.527 (1)} & 0.438 (14) & 0.437 (15) & 0.474 (10) & 0.499 (7) & 0.509 (5) & 0.507 (6) & 0.512 (4) \\
Shuttle & 0.129 (7) & 0.069 (16) & 0.355 (4) & 0.090 (11) & 0.119 (8) & 0.114 (9) & 0.040 (17) & 0.175 (5) & 0.388 (3) & 0.022 (19) & 0.025 (18) & 0.160 (6) & \textbf{0.512 (1)} & 0.112 (10) & 0.084 (13) & 0.071 (15) & 0.090 (12) & 0.079 (14) & 0.436 (2) \\
SpamBase & 0.476 (9) & 0.480 (6) & 0.355 (18) & 0.533 (3) & 0.479 (7) & 0.439 (10) & 0.399 (14) & 0.349 (19) & 0.367 (17) & 0.404 (13) & 0.421 (12) & 0.377 (16) & 0.392 (15) & 0.431 (11) & 0.494 (5) & 0.479 (8) & \textbf{0.559 (1)} & 0.557 (2) & 0.530 (4) \\
Stamps & 0.267 (10) & 0.307 (8) & 0.333 (3) & 0.333 (4) & 0.332 (5) & 0.265 (11) & 0.196 (13) & 0.113 (19) & 0.164 (16) & 0.123 (18) & 0.146 (17) & 0.174 (15) & 0.213 (12) & 0.189 (14) & 0.334 (2) & \textbf{0.345 (1)} & 0.309 (7) & 0.327 (6) & 0.269 (9) \\
WBC & 0.483 (14) & 0.882 (3) & 0.875 (5) & 0.827 (9) & 0.858 (8) & 0.605 (11) & 0.274 (18) & 0.537 (12) & 0.368 (17) & 0.394 (16) & 0.501 (13) & 0.750 (10) & 0.111 (19) & 0.400 (15) & \textbf{0.895 (1)} & 0.877 (4) & 0.874 (7) & 0.885 (2) & 0.875 (6) \\
WDBC & 0.647 (10) & 0.647 (11) & 0.691 (6) & 0.716 (4) & 0.686 (7) & 0.684 (8) & 0.092 (19) & 0.551 (13) & 0.316 (16) & 0.630 (12) & 0.501 (15) & 0.512 (14) & 0.197 (18) & 0.275 (17) & \textbf{0.800 (1)} & 0.678 (9) & 0.760 (3) & 0.694 (5) & 0.768 (2) \\
WPBC & 0.233 (7) & 0.231 (9) & 0.232 (8) & 0.239 (5) & 0.230 (10) & 0.229 (11) & 0.243 (4) & 0.206 (19) & 0.266 (2) & 0.229 (12) & 0.207 (18) & 0.227 (14) & \textbf{0.275 (1)} & 0.216 (17) & 0.265 (3) & 0.227 (15) & 0.229 (13) & 0.225 (16) & 0.235 (6) \\
Waveform & 0.078 (8) & 0.061 (12) & 0.131 (4) & 0.055 (16) & 0.067 (9) & 0.080 (7) & 0.029 (19) & 0.033 (18) & 0.056 (14) & 0.058 (13) & 0.066 (10) & 0.141 (3) & 0.035 (17) & 0.063 (11) & \textbf{0.168 (1)} & 0.056 (15) & 0.115 (6) & 0.118 (5) & 0.166 (2) \\
Wilt & 0.048 (11) & 0.045 (16) & 0.053 (8) & 0.041 (19) & 0.044 (18) & 0.055 (6) & \textbf{0.283 (1)} & 0.089 (3) & 0.046 (14) & 0.054 (7) & 0.052 (9) & 0.083 (4) & 0.147 (2) & 0.049 (10) & 0.079 (5) & 0.045 (17) & 0.048 (12) & 0.046 (15) & 0.048 (13) \\
annthyroid & 0.195 (13) & 0.314 (7) & 0.204 (12) & 0.366 (5) & 0.290 (8) & 0.247 (10) & 0.161 (15) & 0.128 (17) & 0.120 (18) & 0.135 (16) & 0.114 (19) & 0.173 (14) & \textbf{0.655 (1)} & 0.243 (11) & 0.257 (9) & 0.336 (6) & 0.452 (3) & 0.395 (4) & 0.568 (2) \\
arrhythmia & 0.424 (14) & 0.479 (4) & 0.464 (7) & \textbf{0.502 (1)} & 0.444 (11) & 0.431 (13) & 0.313 (15) & 0.257 (19) & 0.309 (16) & 0.460 (8) & 0.450 (10) & 0.442 (12) & 0.291 (17) & 0.267 (18) & 0.459 (9) & 0.482 (2) & 0.474 (6) & 0.480 (3) & 0.475 (5) \\
breastw & 0.895 (9) & 0.969 (4) & 0.392 (18) & 0.967 (5) & \textbf{0.979 (1)} & 0.940 (8) & 0.883 (10) & 0.782 (12) & 0.320 (19) & 0.686 (15) & 0.728 (14) & 0.748 (13) & 0.786 (11) & 0.569 (17) & 0.588 (16) & 0.979 (2) & 0.955 (6) & 0.955 (7) & 0.977 (3) \\
glass & 0.076 (16) & 0.093 (11) & 0.083 (14) & 0.116 (7) & 0.085 (13) & 0.095 (10) & 0.195 (2) & 0.052 (18) & 0.041 (19) & \textbf{0.205 (1)} & 0.154 (5) & 0.171 (4) & 0.172 (3) & 0.078 (15) & 0.125 (6) & 0.116 (8) & 0.092 (12) & 0.104 (9) & 0.058 (17) \\
ionosphere & 0.636 (17) & 0.809 (6) & 0.799 (9) & 0.758 (13) & 0.806 (8) & 0.752 (14) & \textbf{0.949 (1)} & 0.599 (18) & 0.565 (19) & 0.735 (16) & 0.742 (15) & 0.914 (3) & 0.916 (2) & 0.809 (7) & 0.790 (11) & 0.910 (4) & 0.793 (10) & 0.813 (5) & 0.771 (12) \\
letter & 0.140 (7) & 0.087 (15) & 0.244 (4) & 0.088 (12) & 0.134 (8) & 0.225 (5) & 0.311 (2) & 0.094 (10) & 0.081 (19) & 0.100 (9) & 0.085 (16) & \textbf{0.383 (1)} & 0.269 (3) & 0.088 (14) & 0.209 (6) & 0.091 (11) & 0.085 (17) & 0.088 (13) & 0.082 (18) \\
lympho & 0.782 (10) & 0.944 (3) & 0.857 (6) & 0.976 (2) & 0.877 (5) & 0.827 (8) & 0.151 (18) & 0.519 (14) & 0.056 (19) & 0.897 (4) & 0.808 (9) & 0.640 (12) & 0.323 (16) & 0.403 (15) & 0.173 (17) & \textbf{1.000 (1)} & 0.720 (11) & 0.852 (7) & 0.556 (13) \\
mammography & 0.184 (9) & 0.221 (6) & 0.121 (12) & 0.221 (7) & 0.252 (3) & 0.198 (8) & 0.171 (11) & 0.100 (15) & 0.070 (19) & 0.073 (18) & 0.097 (16) & 0.105 (13) & 0.176 (10) & 0.102 (14) & 0.082 (17) & 0.238 (4) & 0.269 (2) & 0.235 (5) & \textbf{0.284 (1)} \\
mnist & 0.246 (15) & 0.265 (13) & 0.379 (5) & 0.262 (14) & 0.274 (12) & 0.295 (9) & 0.393 (3) & 0.225 (17) & 0.207 (18) & 0.392 (4) & 0.333 (8) & 0.350 (7) & 0.412 (2) & \textbf{0.442 (1)} & 0.371 (6) & 0.246 (16) & 0.282 (10) & 0.282 (11) & 0.160 (19) \\
musk & 0.506 (14) & \textbf{1.000 (1)} & 0.090 (17) & 1.000 (2) & 1.000 (3) & 0.818 (11) & 0.450 (15) & 0.856 (10) & 0.043 (18) & 0.998 (8) & 0.440 (16) & 0.025 (19) & 0.517 (13) & 0.688 (12) & 0.994 (9) & 1.000 (4) & 1.000 (5) & 1.000 (6) & 1.000 (7) \\
optdigits & 0.037 (12) & 0.051 (6) & 0.021 (17) & 0.043 (10) & 0.053 (5) & 0.051 (7) & 0.023 (16) & 0.018 (19) & 0.032 (13) & 0.070 (3) & 0.043 (11) & 0.021 (18) & 0.029 (14) & 0.028 (15) & 0.072 (2) & 0.055 (4) & 0.050 (8) & 0.048 (9) & \textbf{0.114 (1)} \\
pendigits & 0.195 (10) & 0.279 (3) & 0.044 (16) & 0.255 (7) & 0.258 (6) & 0.211 (9) & 0.035 (18) & 0.083 (13) & 0.036 (17) & 0.182 (11) & 0.228 (8) & 0.033 (19) & 0.048 (15) & 0.173 (12) & 0.066 (14) & 0.266 (5) & \textbf{0.289 (1)} & 0.272 (4) & 0.285 (2) \\
pima & 0.474 (11) & 0.500 (6) & 0.493 (7) & 0.465 (12) & 0.501 (5) & 0.520 (2) & 0.405 (15) & 0.431 (14) & 0.398 (16) & 0.368 (18) & 0.382 (17) & 0.504 (4) & 0.431 (13) & 0.349 (19) & 0.490 (8) & 0.505 (3) & 0.480 (10) & 0.487 (9) & \textbf{0.522 (1)} \\
satellite & 0.571 (11) & 0.660 (3) & 0.397 (18) & 0.660 (4) & \textbf{0.682 (1)} & 0.609 (7) & 0.508 (15) & 0.574 (10) & 0.479 (16) & 0.527 (14) & 0.587 (8) & 0.332 (19) & 0.580 (9) & 0.540 (12) & 0.415 (17) & 0.658 (5) & 0.658 (6) & 0.671 (2) & 0.535 (13) \\
satimage-2 & 0.709 (11) & 0.926 (5) & 0.142 (16) & 0.915 (8) & 0.944 (3) & 0.729 (10) & 0.100 (17) & 0.394 (12) & 0.063 (18) & 0.961 (2) & 0.889 (9) & 0.033 (19) & 0.142 (15) & 0.380 (14) & 0.393 (13) & 0.921 (7) & 0.927 (4) & 0.924 (6) & \textbf{0.965 (1)} \\
speech & 0.026 (4) & 0.018 (17) & 0.020 (11) & 0.027 (2) & 0.021 (10) & 0.026 (5) & 0.019 (13) & 0.023 (9) & 0.027 (3) & 0.019 (14) & 0.019 (12) & 0.024 (8) & 0.018 (16) & \textbf{0.038 (1)} & 0.019 (15) & 0.018 (18) & 0.026 (6) & 0.026 (7) & 0.017 (19) \\
thyroid & 0.464 (8) & 0.557 (6) & 0.335 (13) & 0.380 (12) & 0.486 (7) & 0.434 (9) & 0.309 (14) & 0.146 (16) & 0.049 (19) & 0.176 (15) & 0.118 (18) & 0.120 (17) & \textbf{0.729 (1)} & 0.565 (5) & 0.410 (10) & 0.612 (3) & 0.651 (2) & 0.604 (4) & 0.406 (11) \\
vertebral & 0.095 (12) & 0.096 (11) & 0.088 (17) & 0.092 (15) & 0.087 (18) & 0.090 (16) & 0.116 (5) & 0.144 (4) & 0.103 (7) & 0.156 (3) & 0.174 (2) & 0.099 (9) & 0.115 (6) & \textbf{0.193 (1)} & 0.000 (19) & 0.098 (10) & 0.094 (13) & 0.093 (14) & 0.099 (8) \\
vowels & 0.169 (9) & 0.138 (13) & 0.385 (4) & 0.106 (15) & 0.228 (7) & 0.192 (8) & 0.458 (3) & 0.067 (17) & 0.057 (19) & 0.140 (12) & 0.063 (18) & \textbf{0.615 (1)} & 0.362 (6) & 0.089 (16) & 0.369 (5) & 0.127 (14) & 0.155 (11) & 0.165 (10) & 0.491 (2) \\
wbc & 0.612 (6) & 0.608 (7) & 0.650 (3) & 0.582 (11) & 0.597 (8) & 0.616 (5) & 0.241 (18) & 0.415 (16) & 0.447 (14) & 0.596 (9) & 0.417 (15) & 0.378 (17) & 0.203 (19) & 0.472 (13) & 0.542 (12) & 0.591 (10) & 0.639 (4) & \textbf{0.668 (1)} & 0.665 (2) \\
wine & 0.246 (6) & 0.213 (12) & 0.290 (2) & 0.234 (8) & 0.253 (5) & 0.272 (3) & 0.093 (17) & 0.112 (15) & 0.081 (18) & 0.220 (9) & 0.108 (16) & 0.239 (7) & 0.256 (4) & 0.123 (14) & 0.000 (19) & 0.211 (13) & 0.215 (10) & 0.215 (11) & \textbf{0.364 (1)} \\
\midrule
\textbf{Average} & 0.337 (10.51) & 0.398 (8.26) & 0.330 (9.41) & 0.392 (8.87) & 0.396 (8.10) & 0.367 (9.13) & 0.276 (12.15) & 0.275 (14.10) & 0.208 (14.59) & 0.347 (10.97) & 0.314 (12.72) & 0.302 (10.69) & 0.314 (10.51) & 0.293 (12.03) & 0.348 (9.03) & 0.408 (8.05) & 0.413 (7.00) & 0.414 (6.97) & \textbf{0.422 (6.90)} \\
\textbf{Std Dev} & 0.243 & 0.308 & 0.248 & 0.302 & 0.300 & 0.264 & 0.223 & 0.227 & 0.171 & 0.285 & 0.255 & 0.242 & 0.231 & 0.209 & 0.267 & 0.313 & 0.294 & 0.297 & 0.283 \\
\bottomrule
\end{tabular}
}
\end{table*}

\begin{figure}[h]
\centering
\includegraphics[width=1.0\linewidth]{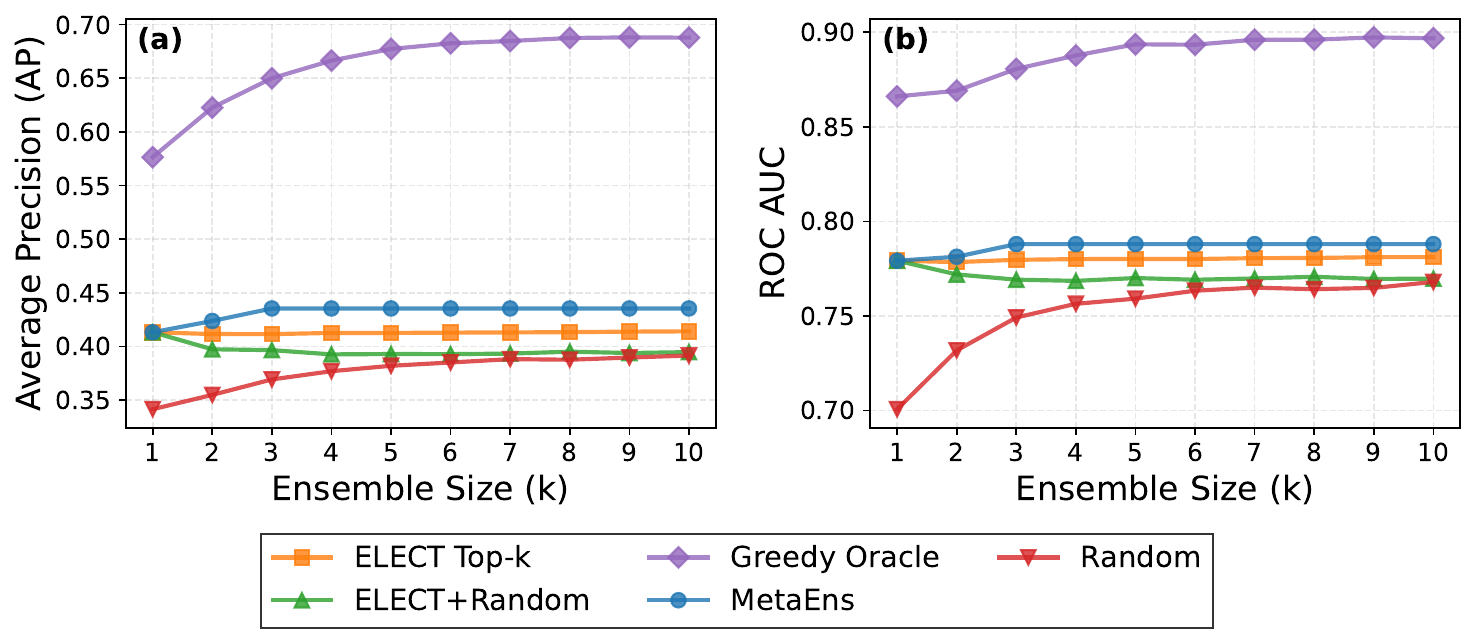}
\caption{Ensemble size impact on performance (AP) and robustness (ROC AUC). \textbf{(Left)} Average Precision vs ensemble size shows that \framework achieves rapid performance gains with small ensembles ($k \leq 3$), then plateaus, confirming the effectiveness of the family-risk regularizer in adaptive ensemble stopping. \textbf{(Right)} ROC AUC demonstrates consistent robustness across different ensemble sizes. Greedy Oracle (with oracle access to test labels) provides an upper bound, while Random selection serves as the lower bound. ELECT performs comparably to single models ($k=1$) across ensemble sizes due to its ranking-based selection strategy.}
\label{fig:ensemble_impact}
\end{figure}

\paragraph{Results Analysis.}
The combined results demonstrate several key findings:

\begin{itemize}
\item \textbf{Scalability:} \framework successfully handles the expanded 310-model pool with minimal performance impact, validating its applicability to larger model collections.
\item \textbf{Robustness:} Performance remains stable across different pool compositions, indicating robust meta-learning that generalizes across model types.
\item \textbf{Discriminative Selection:} Analysis reveals that despite several neural network models achieving competitive individual performance, \framework predominantly selects classical models, demonstrating intelligent cross-dataset pattern recognition.
\item \textbf{Competitive Performance:} \framework achieves the best performance on 10 out of 39 datasets (25.6\%) on the expanded pool, demonstrating consistent competitive behavior across diverse anomaly detection tasks.
\end{itemize}

The stable performance across pool configurations validates \framework's design principle of leveraging comprehensive cross-dataset information for intelligent model selection, supporting its applicability as new model architectures emerge in the anomaly detection landscape.

\subsection{Ensemble Size Impact Analysis}
\label{app:subsec:ensemble_size}

To understand how ensemble size affects performance, we analyze the relationship between the number of models and detection quality. Figure~\ref{fig:ensemble_impact} presents the Average Precision (AP) and ROC AUC as a function of ensemble size for several representative methods.

A key finding from our experiments is that \framework employs adaptive ensemble sizing, selecting an average of 2.2 models per dataset across the 39-dataset benchmark. This adaptive behavior contrasts sharply with fixed-size baselines: \texttt{ELECT} Top-10 always uses 10 models, while Random Ensemble uses a fixed best $k=3$. The ability to adjust ensemble size based on dataset characteristics is a crucial advantage of our approach.

The performance comparison in Table~\ref{tab:baselines} reveals important insights about ensemble construction strategies. \texttt{ELECT} Top-10, despite using 10 models, achieves only 0.4117 AP---a marginal 0.0048 improvement over \texttt{ELECT} (Top-1) with 0.4069 AP. This negligible gain from expanding to 10 models suggests that simple rank-based aggregation without diversity consideration leads to redundant model selection. In contrast, \framework achieves 0.4308 AP with an average of only 2.2 models, demonstrating that carefully selected small ensembles can substantially outperform large ensembles of top-ranked models.

The Random Ensemble baseline with best $k=3$ achieves 0.3759 AP, performing worse than both the single \texttt{ELECT} selector (0.4069 AP) and \framework (0.4308 AP). This indicates that expanding ensembles with randomly chosen partners actively degrades performance by introducing low-quality models that add noise to the ensemble prediction. This finding validates our hypothesis that partner selection must be guided by both quality and diversity considerations.

Our adaptive stopping mechanism enables dataset-specific ensemble construction: simple datasets may benefit from small ensembles to avoid overfitting, while more complex datasets with diverse outlier patterns may require additional models to capture complementary perspectives. Fixed ensemble size baselines cannot adapt to this heterogeneity, leading to suboptimal performance across the benchmark.


\end{document}